\documentclass{article}

\usepackage[preprint]{neurips_2026}

\usepackage[utf8]{inputenc}
\usepackage[T1]{fontenc}
\usepackage{hyperref}
\usepackage{url}
\usepackage{microtype}
\usepackage{graphicx}
\usepackage{nicefrac}
\usepackage{xcolor}
\usepackage{subcaption}
\usepackage{booktabs}
\usepackage{xspace}
\usepackage{algorithm}
\usepackage{algorithmic}

\usepackage{amsmath}
\usepackage{amssymb}
\usepackage{mathtools}
\usepackage{amsthm}

\usepackage[capitalize,noabbrev]{cleveref}

\theoremstyle{plain}

\theoremstyle{definition}

\theoremstyle{remark}

\usepackage[textsize=tiny]{todonotes}

\title{Laya: A LeJEPA Approach to EEG via Latent Prediction over Reconstruction}

\author{%
  Saarang Panchavati \\
  Department of Medical Informatics\\
  UCLA\\
  Los Angeles, USA \\
  \texttt{saarang@ucla.edu} \\
  \And
  Uddhav Panchavati \\
  Department of Cognitive Science\\
  UCSD\\
  San Diego, USA \\
  \And
  Hiroki Nariai \\
  David Geffen School of Medicine\\
  UCLA Mattel Children’s Hospital\\
  Los Angeles, USA \\
  \And
  Corey Arnold \\
  Department of Medical Informatics\\
  UCLA\\
  Los Angeles, USA \\
  \And
  William Speier \\
  Department of Medical Informatics\\
  UCLA\\
  Los Angeles, USA \\
}

\begin{document}

\maketitle

\newcommand{\ours}{\texttt{Laya}\xspace}
\newcommand{\ourss}{\texttt{Laya-S}\xspace}
\newcommand{\oursf}{\texttt{Laya-B}\xspace}

\begin{abstract}
Electroencephalography (EEG) is a widely used tool for studying brain function, with applications in clinical neuroscience, diagnosis, and brain-computer interfaces (BCIs). Recent EEG foundation models trained on large unlabeled corpora aim to learn transferable representations, but their effectiveness remains unclear; reported improvements over smaller task-specific models are often modest, sensitive to downstream adaptation and fine-tuning strategies, and limited under linear probing. We hypothesize that one contributing factor is the reliance on signal reconstruction as the primary self-supervised learning (SSL) objective, which biases representations toward high-variance artifacts rather than task-relevant neural structure. To address this limitation, we explore an SSL paradigm based on Joint Embedding Predictive Architectures (JEPA), which learn by predicting latent representations instead of reconstructing raw signals. We introduce \ours, the first EEG foundation model based on LeJEPA. We show that latent prediction yields representations that encode semantic structure in EEG: \ours embeddings track clinically meaningful state changes such as seizure onset, are resilient to noise, and achieve the strongest mean clinical accuracy under frozen linear probing, with particular gains on tasks where relevant neural patterns are subtle and easily obscured by artifacts. Controlled ablations against matched MAE variants confirm that the choice of pretraining objective, rather than architecture or data, is the primary driver of these gains.

\end{abstract}

\section{Introduction}

Electroencephalography (EEG) is a non-invasive method of recording electrical activity in the brain with good temporal resolution. Due to its low cost and ease of accessibility, it is a critical tool for both clinical neurology and brain-computer interfaces (BCIs). EEG analysis remains challenging due to low signal-to-noise ratio and significant inter-subject variability which hinder generalization and limit the discovery of robust, interpretable biomarkers. Recently, deep learning has emerged as an exciting alternative to manual visualization or hand-crafted pipelines \citep{roy2019deep, craik_deep_2019}. Supervised approaches have achieved strong performance on task-specific datasets, but still fail to generalize to new subjects or recording setups.

Recent years have seen growing interest in EEG foundation models — large neural networks pretrained on large EEG corpora to learn transferable representations across tasks and datasets \citep{kuruppu_eeg_2025}.

Models such as LaBraM \citep{Jiang2024LaBraM}, LUNA \citep{Doner2025LUNA}, CBraMod \citep{wang2025cbramod}, and REVE \citep{ElOuahidi2025REVE} have demonstrated state-of-the-art capabilities in both clinical applications and BCI tasks like motor imagery decoding. By leveraging large, cross-subject EEG corpora, these approaches highlight the potential of EEG foundation models to learn reusable neural representations that generalize across both long-duration, state-level clinical tasks, and short-horizon, event-driven BCI tasks.

Despite these successes, however, recent benchmarking studies such as EEG-Bench~\citep{Kastrati2025EEGBench}  indicate that EEG foundation models do not consistently outperform simpler baselines. Reported gains are often modest and highly task-dependent. Frozen linear probing often yields near-chance performance, indicating weak intrinsic representations that require extensive adaptation and finetuning to be competitive. This raises questions about the usefulness of these models in the real world, particularly given the substantial computational cost of pretraining~\citep{yang_are_2026,Kastrati2025EEGBench,wu2025adabrainbenchbenchmarkingbrainfoundation}.

Several recurring limitations help explain these mixed results. Generalization across tasks and datasets remains limited, with pretrained representations often failing to transfer beyond the pretraining distribution. Moreover, scaling laws in EEG are not yet well-established; larger models do not consistently yield better transfer performance, especially under frozen evaluation protocols \citep{yang_are_2026, kuruppu_eeg_2025}. Almost all existing approaches rely on reconstruction-based pretraining objectives applied directly to raw EEG signals. Because EEG is inherently noisy and dominated by non-task-relevant variability, optimizing for signal-space reconstruction may encourage models to learn nuisance variables rather than semantic structure \citep{littwin_how_2024}, resulting in weak linear-probe performance and heavy reliance on downstream fine-tuning.

These observations suggest that while EEG foundation models are a promising direction, current approaches leave significant room for improvement. In particular, there is a need for representation learning objectives that better align with downstream discriminative tasks, improve data efficiency, and generalize across heterogeneous EEG domains without extensive task-specific adaptation.

\subsection{JEPA and LeJEPA}

Joint Embedding Predictive Architectures (JEPAs) \citep{lecun2022path} learn representations by predicting latent embeddings rather than reconstructing raw inputs, encouraging semantically rich features while discarding unpredictable variation \citep{assel_joint_2025, balestriero_learning_2024, littwin_how_2024}. A key challenge is preventing representation collapse; LeJEPA \citep{balestriero2025lejepa} addresses this through explicit geometric regularization (SIGReg), removing reliance on momentum-updated target encoders.

This is particularly relevant for EEG, where high-variance artifacts often dominate the signal. We hypothesize that latent prediction objectives naturally filter irrelevant noise, yielding better EEG representations. While emerging works have explored JEPA-style objectives for EEG \citep{hojjati2025videoeegadaptingjoint, guetschel2024s-jepa}, they rely on standard heuristics to prevent collapse and have not demonstrated broad generalization at foundation model scale. We adapt LeJEPA to a masked temporal prediction setting, eliminating asymmetric encoders and learning by predicting temporally masked representations \citep{bardes2024vjepa, assran2025vjepa2}. Our results suggest that latent prediction objectives are a promising direction for large-scale EEG representation learning, with the clearest benefits in noisy clinical tasks where meaningful state-dependent structure can be subtle.

\subsection{Our Contributions}

\begin{itemize}
\item We introduce Laya, the first LeJEPA-based EEG foundation model, designed for clinically realistic settings with scarce labels, heterogeneous montages, and substantial noise.
\item We show that \ours learns semantically organized representations: embeddings track seizure onset and distinguish imagined left- from right-hand movements as spatially distinct structures, in both cases more clearly than reconstruction-based baselines---indicating that latent prediction induces brain-state-aligned organization across clinical and BCI domains.
\item We demonstrate that this representational structure transfers to noisy clinical EEG tasks: latent prediction achieves the strongest clinical mean accuracy under frozen linear probing.
\item We show that \ours learns representations that are resilient to noise—maintaining performance at SNR levels where reconstruction-based models degrade substantially.
\item We confirm via controlled MAE ablations that the pretraining objective — not architecture or data — is the primary driver of these clinical gains.
\item We extend EEG-Bench with noise robustness evaluation, linear probing protocols, and additional baselines including LUNA, CBraMod, and REVE.
\end{itemize}

\section{Related Works}
\subsection{JEPAs}

Joint Embedding Predictive Architectures (JEPAs) learn by predicting latent representations rather than reconstructing pixels or raw inputs. I-JEPA and V-JEPA demonstrate this principle for images and video using masked latent prediction with target encoders~\citep{assran2023ijepa,bardes2024vjepa}. LeJEPA augments the latent-prediction objective with Sketched Isotropic Gaussian Regularization (SIGReg), removing reliance on momentum-updated EMA teachers while maintaining stable, collapse-free pretraining~\citep{balestriero2025lejepa}.

\subsection{EEG Foundation Models}

Recent EEG foundation models adapt masked modeling and transformer architectures to large-scale EEG pretraining. LaBraM tokenizes EEG patches for masked reconstruction, CBraMod separates spatial and temporal dependencies with criss-cross attention, REVE uses coordinate-based spatio-temporal encodings for flexible electrode layouts, and LUNA compresses variable-channel EEG into learned latent queries~\citep{Jiang2024LaBraM,wang2025cbramod,ElOuahidi2025REVE,Doner2025LUNA}. These models establish strong baselines for transfer, but rely primarily on reconstruction objectives; our work instead evaluates latent prediction as the central pretraining signal.

\subsection{JEPA-inspired EEG Foundation models}

Recent work has explored JEPA-style training at scale and across modalities, including spatial block masking over EEG channels, latent prediction over fMRI regions, and video-inspired spatiotemporal masking for EEG~\citep{guetschel2024s-jepa,dong2024brainjepa,hojjati2025videoeegadaptingjoint}. EEG-DINO instead uses hierarchical self-distillation rather than masking-based prediction, but is evaluated on a narrower downstream suite~\citep{wang2025eegdino}. \ours differs by pairing a LeJEPA objective with topology-agnostic EEG modeling and evaluating frozen representations across a broader clinical and BCI benchmark.

\section{Methods}

\subsection{LeJEPA}

Joint-Embedding Predictive Architectures (JEPA) learn representations by predicting latent embeddings of masked or augmented views rather than reconstructing raw inputs. Formally, let $x$ be an input split into a context $x_C$ and a target $x_T$. Let $f_\theta:\mathcal{X}\rightarrow\mathbb{R}^d$ be an encoder and $g_\phi:\mathbb{R}^d\rightarrow\mathbb{R}^d$ a latent predictor. The JEPA objective minimizes:
\begin{equation}
\mathcal{L}_{\text{JEPA}}
=
\mathbb{E}_x\,\| g_\phi(f_\theta(x_C)) - f_\theta(x_T) \|_2^2.
\end{equation}
A key challenge is preventing representation collapse; prior approaches address this with momentum-updated target encoders, which complicate optimization. LeJEPA \citep{balestriero2025lejepa} introduces \emph{Sketched Isotropic Gaussian Regularization} (SIGReg), which stabilizes latent prediction by encouraging batch-level embeddings to be diverse and approximately isotropic, removing reliance on EMA teachers. The combined objective is:
\begin{equation}
\mathcal{L}_{\text{LeJEPA}}
=
\mathcal{L}_{\text{JEPA}}
+
\lambda\,\mathcal{L}_{\text{SIGReg}},
\end{equation}
where $\mathcal{L}_{\text{SIGReg}}$ operates directly on batches of latent representations to prevent collapse and enable stable representation learning that translates directly to downstream performance.
 
\subsection{\ours}

\begin{figure*}[t]
  \begin{center}
    \centerline{\includegraphics[width=.8\textwidth]{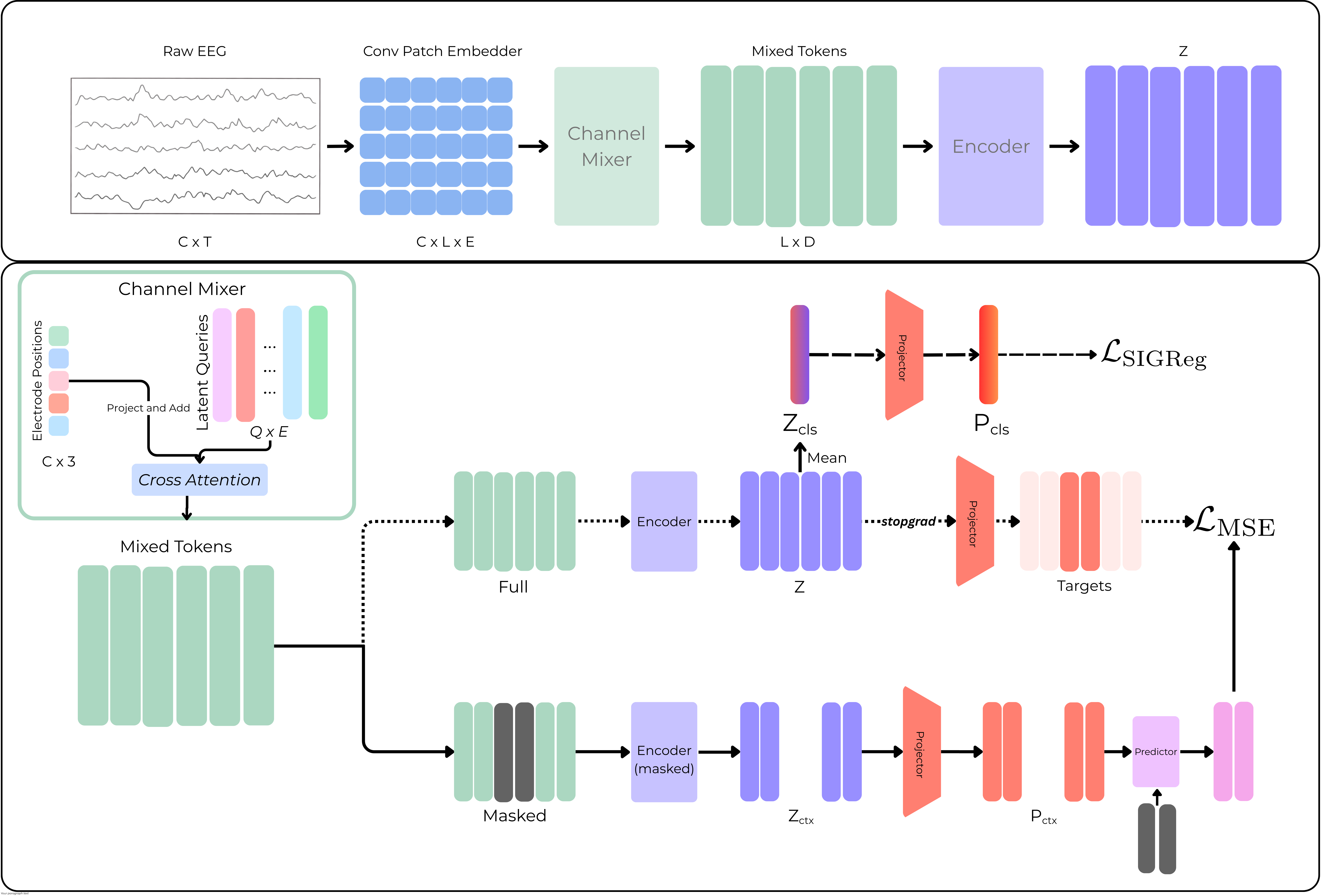}}
        \caption{\textbf{Laya architecture overview.} Raw EEG is embedded into temporal patches, compressed across channels by a learned channel mixer, and encoded into latent brain-state representations. During pretraining, the model predicts masked latent targets from visible context while SIGReg regularizes the embedding geometry.}
    \label{fig:laya}
  \end{center}
\end{figure*}

\subsubsection{Pretraining Objective}

We adopt two components from LeJEPA: the JEPA latent prediction objective (Eq.~1) and SIGReg for collapse prevention. We depart from the original formulation in two ways. First, rather than using augmented views as context and target, we instantiate $x_C$ and $x_T$ as observed and masked temporal segments of the same EEG sequence---a masked prediction paradigm analogous to V-JEPA \citep{bardes2024vjepa} and S-JEPA \citep{guetschel2024s-jepa}. Second, we do not employ a cross-view invariance loss; SIGReg is applied solely as a geometric regularizer on the global embedding of the masked sequence.
Given EEG input $\mathbf{x}\in \mathbb{R}^{B \times C \times T}$, we extract and embed non-overlapping temporal patches to obtain $\tilde{\mathbf{x}} \in \mathbb{R}^{B \times C \times N \times E}$, where $N$ is the number of patches, and $E$ is the latent dimension of the patch embedder. We then apply a learned channel mixer to produce a sequence $\mathbf{S} \in \mathbb{R}^{B \times N \times D}$ of latent brain states which we interpret as mixed-channel representations at each temporal position. We mask contiguous blocks of time and task the model with predicting these masked regions in latent space. The full pretraining step is provided in Appendix~\cref{app:pretraining-algorithm}.


\subsubsection{Patch Embedder}
\label{sec:patchembedder}

We use a 1D convolution to convert raw EEG time series $\mathbf{x}\in \mathbb{R}^{B \times C \times T}$ into patch-level embeddings $\tilde{\mathbf{x}}\in \mathbb{R}^{B \times C \times N \times E}$. A depthwise 1D convolution is applied independently to each channel, with kernel size and stride equal to the patch length \(P\), producing non-overlapping temporal patches. Each convolutional kernel maps \(P\) contiguous time samples directly to an \(E\)-dimensional embedding. By treating channels independently at this stage, the patch embedder avoids imposing inductive biases tied to specific electrode montages.

 \subsubsection{Dynamic Channel Mixer}



To aggregate information across EEG channels, we adopt the channel-mixing mechanism introduced in LUNA \citep{Doner2025LUNA}, which uses learned queries and cross-attention to compress multi-channel EEG into a compact latent representation. At each temporal position, learned query vectors attend across channels, enabling the model to summarize spatially distributed activity into a fixed-dimensional embedding. To make this operation spatially aware yet montage-agnostic, electrode coordinates are encoded and incorporated into the channel representations prior to attention, allowing the mixer to exploit spatial relationships without assuming a fixed electrode layout.

Given patch embeddings $\tilde{\mathbf{x}}\in\mathbb{R}^{B\times C\times N\times E}$ and electrode coordinates $\mathbf{E}\in\mathbb{R}^{C\times 3}$, we incorporate spatial information by encoding electrode coordinates with fixed Fourier features and adding the resulting embeddings to the channel representations. Learned query vectors then attend over these spatially augmented channel embeddings at each temporal patch, producing
\[
\mathbf{A} \in \mathbb{R}^{B \times N \times N_{\text{q}} \times E},
\]
where $N_{\text{q}}$ is the number of queries. We flatten the final two dimensions of $\mathbf{A}$ and apply a linear projection to obtain a channel-mixed latent brain-state sequence
\[
\mathbf{S}\in\mathbb{R}^{B\times N\times D}.
\]
By explicitly encoding spatial information and using multiple learned query outputs per patch, the channel mixer produces a rich, montage-agnostic representation that decouples spatial aggregation from temporal modeling.

\paragraph{Query Specialization Loss.}
Following LUNA, we regularize the query--channel affinity induced by the channel mixer to encourage specialization across learned queries. Let $\mathbf{W}\in\mathbb{R}^{B\times N_{\text{q}}\times C}$ denote the query--channel affinity matrix obtained by averaging attention weights across heads. We penalize pairwise similarity between different query affinity vectors by minimizing the off-diagonal entries of $\mathbf{W}\mathbf{W}^T$.
This loss discourages redundant queries while preserving soft spatial aggregation.

\subsubsection{Pretraining}

Hyperparameters and implementation details are provided in Appendix \ref{sec:pretraining-details}.

\paragraph{Masking Strategy.}
Consistent with previous EEG foundation model approaches, and other video modeling \citep{Jiang2024LaBraM, tong_videomae_2022}, we employ a block masking strategy that masks contiguous spans of time rather than random patch masking. Block masking forces the model to bridge longer temporal gaps, learning structural dependencies rather than autocorrelation. We mask 60\% of the sequence using randomly sampled blocks spanning 500ms to 1s.

\textbf{Encoder.}
The encoder is a standard transformer encoder implemented using the \href{https://github.com/lucidrains/x-transformers}{x-transformers} library. We use rotary positional embeddings (RoPE) \citep{su_roformer_2023} and do not introduce additional absolute positional embeddings to avoid overfitting to fixed temporal locations under masking.

The encoder is applied in two passes: first over the full sequence to produce embeddings $\mathbf{Z}$ and global summary $\mathbf{z}_{\mathrm{cls}}$ (with \textit{StopGrad} applied to $\mathbf{Z}$), then over the context-only sequence with target positions masked to produce context embeddings $\mathbf{Z}_{\mathrm{ctx}}$.

\textbf{Projector.}
Following \citet{balestriero2025lejepa}, the projector is a 3-layer MLP with batch normalization that expands the representation dimension ($D_{up} \gg D$) before compressing them down to the projector dimension $D_{proj} < D$. The projector is applied to $\mathbf{Z}$, $\mathbf{z}_{\mathrm{cls}}$, and $\mathbf{Z}_{\mathrm{ctx}}$. 

\textbf{Predictor.}
Following MAE and V-JEPA design principles, the predictor is a lightweight Transformer encoder operating at dimension $D_{proj}$. Given the bottlenecked context embeddings $\mathbf{P}_{\mathrm{ctx}}$, it predicts latent representations $\hat{\mathbf{T}}$ at masked temporal positions. Learnable mask tokens are inserted at masked positions before being passed to the predictor. As in the encoder, RoPE is used to encode relative temporal structure.

\textbf{Objective.}
The full \ours training objective extends LeJEPA with the query specialization loss from the channel mixer:
\begin{equation}
\mathcal{L}_{\text{\ours}} = \mathcal{L}_{\text{LeJEPA}} + \gamma\,\mathcal{L}_{\text{query}},
\end{equation}
where $\mathcal{L}_{\text{LeJEPA}}$ minimizes MSE between predicted masked embeddings $\hat{\mathbf{T}}$ and stopped-gradient targets $\mathbf{T}$ at masked positions while applying SIGReg to $\mathbf{p}_{\mathrm{cls}}$, and $\mathcal{L}_{\text{query}}$ encourages specialization across channel mixer queries. In our experiments, $\gamma=1$.

\section{Experiments}

\textbf{Pretraining datasets.}
\ours is pretrained on a diverse corpus of EEG recordings drawn from multiple large-scale and task-specific sources: the Temple University Hospital EEG Corpus \citep{obeid2016tuh}, NMT \citep{khan_nmt_2022}, \href{https://eegdash.org/}{EEGDash} (a filtered aggregation of OpenNeuro EEG datasets), the Healthy Brain Network (HBN) \citep{shirazi2024hbn}, and several MOABB BCI datasets \citep{chevallier_largest_2024}. All datasets are converted into fixed-length chunks using a consistent preprocessing pipeline. After filtering and quality control, the final pretraining corpus contains 913,314 samples totaling 29,109 hours of EEG, spanning 20,940 subjects and 17 distinct channel topologies. This mixture spans a wide range of tasks, montages, and acquisition settings. To prevent leakage into downstream evaluation, we exclude datasets used in EEG-Bench evaluation and remove TUH subjects appearing in downstream validation or test sets; further details are provided in Appendix~\cref{seec:pretraining-data}.

\textbf{Implementation Details.} Both variants use an effective sample batch size of 512, distributed across two NVIDIA L40S GPUs. Input sequences are generated by random sampling 16-second crops from the available 120-second recording segments. Comprehensive hyperparameters and optimization details are provided in \cref{sec:pretraining-details}. We present two variants of \ours: \ourss, and \oursf. \ourss was trained on 10\% (3K hours) of the total training dataset for 10K steps (sampled equally from all datasets in the mix). \oursf was trained on all 30K hours for 20K steps.

\subsection{Downstream Evaluation with EEG-Bench}
\label{sec:eegbench}

\textbf{Benchmark Framework and Baselines.} We evaluate downstream performance using EEG-Bench \citep{Kastrati2025EEGBench}, a standardized benchmarking framework spanning 14 diverse brain--computer interface (BCI) and clinical EEG tasks. We compare \ours against established EEG foundation models: LaBraM \citep{Jiang2024LaBraM}, LUNA \citep{Doner2025LUNA}, CBraMod \citep{wang2025cbramod}, and REVE \citep{ElOuahidi2025REVE}. For all baselines, we use publicly released pretrained checkpoints. Dataset details are provided in \cref{sec:downstream-data}.

\textbf{Fine-tuning protocols.}
We primarily assess representation quality under a linear probing protocol. The pretrained encoder parameters are frozen, and a task-specific linear classification head is trained on the embeddings. 
We focus on frozen linear probing because our goal is to evaluate the intrinsic quality of the pretrained representations while minimizing confounds from model-specific downstream architectures and hyperparameter tuning.


\textbf{Metrics.} We report Balanced Accuracy from linear probing as the primary task-level metric to account for the significant class imbalance common in clinical EEG datasets. For multi-label tasks, metrics are computed on flattened predictions across temporal segments.

\subsection{\ours learns semantically organized representations}
Before evaluating downstream accuracy, we first ask whether latent prediction changes the structure of the learned EEG representations. If \ours is learning clinically meaningful state information rather than reconstructing high-variance signal detail, its embeddings should vary coherently with changes in brain state.

\begin{figure}[ht]
  \begin{center}
  \centerline{\includegraphics[width=\columnwidth, keepaspectratio]{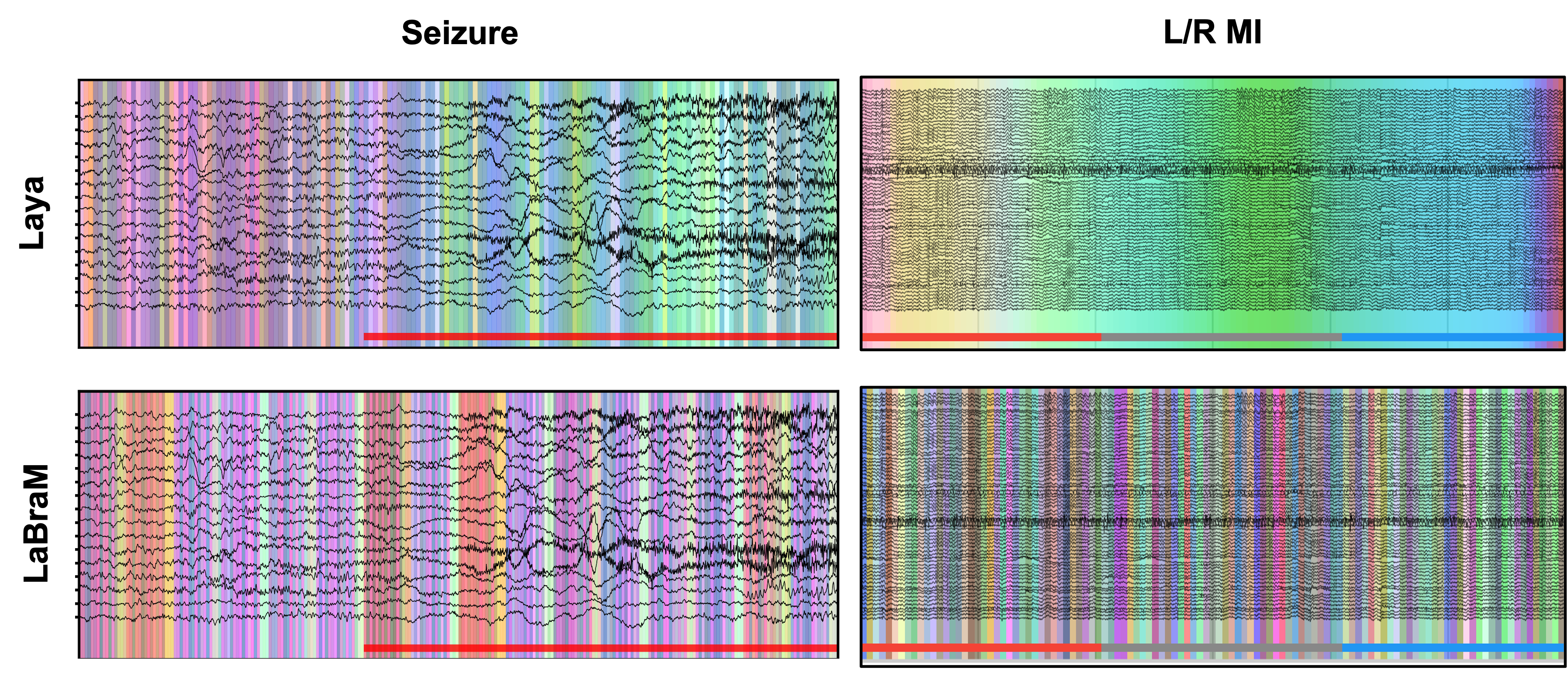}}
    \caption{
    PCA visualization of patch embeddings (three principal components mapped to RGB) on a seizure recording (left) and an L/R motor imagery recording (right). Top row: \oursf; bottom row: LaBraM. On the seizure recording, the red bar marks seizure onset. On the motor imagery recording, color bars indicate motor state: red = right hand, blue = left hand, grey = rest. \oursf shows a clear representational shift at seizure onset and distinct spatial structure across motor states; LaBraM does not.
    }
    \label{pca}
  \end{center}
  \vspace{-10pt}
\end{figure}

To assess learned representations, we visualize patch embeddings from a seizure recording using PCA, following \citet{oquab_dinov2_2024}. We project patch embeddings to three principal components and map them to RGB channels, overlaying the result on the raw EEG (\cref{pca}).

\ours's embeddings show clear temporal structure aligned with seizure onset and termination: at a state change, the colors shift distinctly. In contrast, LaBraM's embeddings fluctuate throughout the recording without clear correspondence to the clinical event. This representational structure generalizes beyond seizures: \cref{pca} (right) shows that \ours also produces distinct embeddings for imagined left- versus right-hand movements, with visually separable color regions across motor states. LaBraM shows no such structure in either setting. This suggests \ours learns features organized around semantically meaningful brain states rather than surface-level signal statistics---a property that holds across both clinical and BCI domains. Further examples can be found in Appendix \ref{app:pca}.

\subsection{Downstream Results on EEG-Bench}
We next evaluate whether this representational structure translates into downstream transfer under frozen linear probing, particularly on noisy clinical EEG tasks where relevant state-dependent patterns can be subtle.

\begin{table}[t]
\centering
\caption{Linear probe performance on motor imagery tasks (balanced accuracy, mean $\pm$ std over 5 seeds). \textbf{Bold}: best, \underline{underline}: second best.}
\label{tab:bci_results}
\resizebox{\columnwidth}{!}{%
\begin{tabular}{lcccccc}
\toprule
Task & LaBraM & LUNA & CBraMod & REVE & \ourss & \oursf \\
\midrule
5-Finger MI & 0.208$\pm$0.005 & 0.202$\pm$0.005 & 0.204$\pm$0.005 & \textbf{0.234}$\pm$0.004 & \underline{0.210}$\pm$0.007 & 0.205$\pm$0.005 \\
LH vs RH MI & 0.471$\pm$0.019 & 0.493$\pm$0.007 & 0.491$\pm$0.015 & \textbf{0.513}$\pm$0.017 & \underline{0.510}$\pm$0.003 & 0.507$\pm$0.004 \\
4-Class MI & \textbf{0.289}$\pm$0.005 & 0.259$\pm$0.009 & 0.271$\pm$0.017 & 0.280$\pm$0.007 & 0.273$\pm$0.005 & \underline{0.288}$\pm$0.007 \\
RH vs Feet MI & 0.524$\pm$0.017 & 0.501$\pm$0.002 & 0.515$\pm$0.011 & 0.508$\pm$0.009 & \textbf{0.586}$\pm$0.002 & \underline{0.578}$\pm$0.002 \\
\midrule
Mean & 0.373 & 0.364 & 0.370 & 0.384 & \textbf{0.395} & \underline{0.394} \\
\bottomrule
\end{tabular}
}
\end{table}

\begin{table}[t]
\centering
\caption{Linear probe performance on clinical tasks (balanced accuracy, mean $\pm$ std over 5 seeds). \textbf{Bold}: best, \underline{underline}: second best.}
\label{tab:clinical_results}
\resizebox{\columnwidth}{!}{%
\begin{tabular}{lcccccc}
\toprule
Task & LaBraM & LUNA & CBraMod & REVE & \ourss & \oursf \\
\midrule
Abnormal & \underline{0.781}$\pm$0.013 & 0.733$\pm$0.018 & 0.605$\pm$0.011 & 0.500$\pm$0.000 & \textbf{0.798}$\pm$0.007 & 0.778$\pm$0.006 \\
Artifact (Binary) & \underline{0.725}$\pm$0.003 & 0.707$\pm$0.023 & 0.690$\pm$0.001 & 0.720$\pm$0.003 & 0.716$\pm$0.001 & \textbf{0.737}$\pm$0.001 \\
Epilepsy & \textbf{0.696}$\pm$0.019 & 0.647$\pm$0.128 & 0.557$\pm$0.017 & \underline{0.667}$\pm$0.018 & 0.657$\pm$0.023 & 0.596$\pm$0.012 \\
mTBI & 0.549$\pm$0.133 & 0.613$\pm$0.052 & 0.499$\pm$0.002 & 0.682$\pm$0.053 & \underline{0.699}$\pm$0.030 & \textbf{0.815}$\pm$0.048 \\
Artifact (Multiclass) & 0.374$\pm$0.006 & 0.288$\pm$0.019 & 0.298$\pm$0.004 & \textbf{0.431}$\pm$0.058 & 0.375$\pm$0.002 & \underline{0.382}$\pm$0.008 \\
OCD & \textbf{0.697}$\pm$0.175 & 0.546$\pm$0.063 & 0.600$\pm$0.173 & \underline{0.633}$\pm$0.000 & 0.627$\pm$0.144 & 0.567$\pm$0.037 \\
Parkinson's & 0.537$\pm$0.139 & 0.500$\pm$0.000 & 0.508$\pm$0.039 & 0.562$\pm$0.031 & \textbf{0.708}$\pm$0.012 & \underline{0.683}$\pm$0.012 \\
Schizophrenia & \textbf{0.562}$\pm$0.080 & 0.423$\pm$0.085 & 0.481$\pm$0.043 & 0.444$\pm$0.100 & 0.479$\pm$0.060 & \underline{0.502}$\pm$0.042 \\
Seizure & 0.678$\pm$0.002 & \underline{0.759}$\pm$0.008 & 0.699$\pm$0.002 & 0.561$\pm$0.023 & \textbf{0.786}$\pm$0.014 & 0.723$\pm$0.005 \\
Sleep Stages & 0.184$\pm$0.000 & 0.182$\pm$0.002 & 0.173$\pm$0.000 & \textbf{0.267}$\pm$0.004 & \underline{0.185}$\pm$0.000 & 0.183$\pm$0.001 \\
\midrule
Mean & 0.578 & 0.540 & 0.511 & 0.547 & \textbf{0.603} & \underline{0.597} \\
\bottomrule
\end{tabular}
}
\end{table}

On motor imagery tasks, performance is mixed across models. \ours achieves the highest mean BCI accuracy (0.395) but does not consistently outperform across individual tasks.

On clinical tasks, \ourss achieves the highest mean accuracy (0.603) despite using only 10\% of pretraining data, outperforming LaBraM (0.578), REVE (0.547), LUNA (0.540), and CBraMod (0.511). Gains are most pronounced on tasks where relevant neural patterns are subtle and easily obscured by high-variance artifacts. Notably, \ourss achieves this with only 10\% of pretraining data, highlighting the sample efficiency of latent prediction-based pretraining.

\oursf leads on binary artifact detection (0.737). On multiclass artifact classification, REVE achieves the highest score (0.431), with \oursf close behind (0.382), while LaBraM, LUNA, and CBraMod trail substantially (0.374, 0.288, 0.298).

\subsection{\ours learns robust representations}
We evaluated robustness by injecting noise (Gaussian, 1/f, EMG, channel dropout) at varying SNR levels with frozen probes. \ours retains 91\% of its clean accuracy at 10 dB SNR on abnormal detection while LaBraM drops to 78\%, consistent with our hypothesis that reconstruction-based models learn features correlated with high-variance input components, making them vulnerable when noise is introduced. Full degradation curves across clinical tasks are shown in Appendix~\cref{app:noise-robust}.

\subsection{Ablations}
\textbf{SIGReg is essential to prevent collapse.} We found that SIGReg regularization is essential for stable training. With $\lambda < 0.02$ the model collapses to a constant embedding across the batch, resulting in chance-level downstream performance. Similarly, removing the projector severely degraded downstream performance, yielding a mean accuracy of 0.476.

\textbf{Latent prediction outperforms reconstruction on clinical tasks.}
To directly test whether the pretraining objective drives clinical gains, we trained two MAE variants using identical architecture, data, and training configuration as \ourss. \texttt{Laya-MAE-Decoder} adds a lightweight decoder to reconstruct raw EEG signals; \texttt{Laya-MAE} predicts the channel-mixed token sequence ($\mathbf{S}$, the output of the channel mixer) rather than the encoder's contextualized representations ($\mathbf{Z}$), isolating the prediction target while keeping all other components fixed.

\begin{table}[t]
\centering
\caption{Controlled MAE ablation. All variants use identical architecture, data, and training configuration. Clinical Mean is averaged over clinical tasks in EEG-Bench. \textbf{Bold}: best.}
\label{tab:mae_ablation}
\resizebox{\columnwidth}{!}{%
\begin{tabular}{llcc}
\toprule
Variant & Objective & BCI Mean & Clinical Mean \\
\midrule
\texttt{Laya-MAE-Decoder} & Raw signal reconstruction & 0.390 & 0.529 \\
\texttt{Laya-MAE} & Channel-mixed token prediction & 0.389 & 0.567 \\
\ourss & Latent prediction & \textbf{0.395} & \textbf{0.603} \\
\oursf & Latent prediction & 0.394 & 0.597 \\
\bottomrule
\end{tabular}
}
\end{table}

Across both BCI and clinical tasks, both MAE variants underperform \ourss and \oursf. \texttt{Laya-MAE-Decoder} trails \ourss by 7.4 points on clinical mean and also underperforms on BCI. These results directly isolate the pretraining objective as the primary driver of \ours's advantage.

\textbf{StopGrad stabilizes longer training.} In Laya, StopGrad serves a narrower role than in I-JEPA or V-JEPA: we use a single shared encoder with no separate teacher network, so StopGrad only decouples target representations from the predictor gradient path while SIGReg prevents dimensional collapse. Removing StopGrad has a small effect at the \ourss scale but substantially degrades clinical performance at 20K steps, indicating that StopGrad becomes increasingly important for preserving representation quality during longer training (Appendix~\cref{app:stopgrad-ablation}).

\section{Discussion}

We introduce \ours, the first application of LeJEPA to EEG representation learning, and show that stable latent prediction enabled by LeJEPA and SIGReg provides a strong alternative to reconstruction-based pretraining for noisy clinical EEG tasks with subtle state-dependent structure. By predicting in latent space, \ours achieves greater sample efficiency, robustness, and semantic organization in these settings.

Across downstream benchmarks, \ours matches or exceeds reconstruction-based baselines. On BCI motor imagery tasks, \ours performs comparably to reconstruction-based baselines in aggregate; however, PCA visualizations reveal that \ours does learn qualitatively distinct representations across left- and right-hand motor states, while LaBraM shows no such structure, suggesting that quantitative parity may reflect task-specific confounds (e.g., subject variability, trial-level decoding) rather than a failure to encode motor-relevant information. In contrast, \ours achieves substantial gains on clinical tasks. \ourss trained on only 10\% of the pretraining data attains the highest mean balanced accuracy. Notably, \ours exhibits strong sample efficiency and learns noise-resilient representations.

Ablation results show that removing SIGReg makes training unstable, underscoring the necessity of explicit geometric regularization. Controlled MAE ablations further confirm that the latent predictive objective — not architecture or data — is the primary driver of Laya's clinical gains.

While the pretraining objective plays a central role, these results also emphasize the importance of architectural inductive biases for EEG, particularly channel mixing. Structured channel interactions are critical for enabling JEPA-style objectives to learn stable and transferable representations, suggesting that objective and architecture must be designed jointly.

\textbf{Conclusion.}
These results indicate that reconstruction is not a necessary (and may be a suboptimal) objective for noisy clinical EEG representation learning. Controlled MAE ablations confirm that this clinical advantage stems directly from the latent predictive objective, rather than architectural or data-driven factors. Latent prediction therefore offers a strong alternative to signal reconstruction for scalable and clinically relevant EEG representation learning, especially when useful features are subtle, noisy, and expensive to annotate at scale.

\subsection{Limitations}
\ourss was only trained on 10\% of the overall curated data, and while this yielded strong results, \oursf did not consistently outperform \ourss on clinical tasks despite $10\times$ more pretraining data and $2\times$ more training steps. We reduced the SIGReg weight from $\lambda = 0.05$ to $\lambda = 0.02$ for Laya-B to maintain stability over longer training, but a systematic sweep over schedule length, $\lambda$, and data mixture was not feasible with our available compute (2 GPUs); we leave a careful study of scaling dynamics for latent-prediction EEG models to future work. Despite this, Laya-B matches or exceeds all reconstruction-based baselines on clinical mean (0.597), suggesting that the latent predictive objective provides a strong performance floor even without optimal scaling.

EEG-Bench focuses on classification tasks and does not extend to retrieval-style tasks like cognitive decoding, emotion recognition, or forecasting (e.g., seizure prediction rather than detection). Evaluating on other benchmarks like AdaBrain-Bench \citep{wu2025adabrainbenchbenchmarkingbrainfoundation} or EEG-FM-Bench \citep{xiong_eeg-fm-bench_2025} is necessary to fully characterize representation quality. Additionally, linear probing may underestimate the utility of learned representations for tasks requiring nonlinear adaptation or subject-specific calibration.
BCI performance remains limited across all self-supervised methods, consistent with recent benchmarks showing that EEG foundation models often struggle to outperform simpler or task-specific decoders on fine-grained motor decoding tasks~\citep{yang_are_2026,Kastrati2025EEGBench}. This suggests that frozen population-level EEG representations may be insufficient for granular task-related changes, and that future approaches may need to model neural activity and behavior jointly, or incorporate subject-level adaptation, to capture the structure required for BCI.

\subsection{Future Work}
Our work suggests that latent prediction can yield semantically meaningful EEG representations, but it also raises several open questions. A key direction is developing a clearer understanding of what structure these representations encode. Further analysis is needed to characterize how temporal, spectral, and spatial information are organized in LeJEPA embeddings, and how this organization differs from that induced by reconstruction-based objectives. \ours's quantitative and qualitative results on seizure-related tasks, together with qualitative evidence of structured motor-state representations, suggest promising applications of JEPA-style architectures for seizure forecasting, interpretable clinical decision support, and BCI systems requiring robust population-level representations, where robustness and semantic organization are critical.

More broadly, our findings suggest that future work on EEG SSL should prioritize principled regularization and objective design over architectural complexity alone.

From a practical perspective, a more systematic search over pretraining recipe hyperparameters — schedule length, SIGReg strength, and data mixture — remains an important next step. Related questions around context length and long-range temporal modeling also remain open.

Progress in EEG foundation models also depends on more extensive and standardized benchmarks, alongside improved data curation and shared pretraining pipelines. Greater agreement on task definitions, evaluation protocols, and preprocessing would enable more meaningful comparisons and faster iteration across the field.


\bibliography{example_paper}
\bibliographystyle{abbrvnat}

\clearpage

\appendix
\section{Pretraining Details}
\label{sec:pretraining-details}

\begin{table}[H]
\centering
\caption{Pretraining details for \ourss (10\% data) and \oursf (full data).}
\label{tab:pretraining_details}
\small
\setlength{\tabcolsep}{5pt}
\begin{tabular}{lll}
\toprule
\textbf{ } & \textbf{\ourss} & \textbf{\oursf} \\
\midrule
\multicolumn{3}{l}{\textbf{Data}} \\
Training data fraction & 10\% & 100\% \\
Max input duration & 16 s (160 timesteps) & 16 s (160 timesteps) \\
Batch size & 256 & 256 \\
\midrule
\multicolumn{3}{l}{\textbf{Model}} \\
Encoder dim & 384 & 384 \\
Encoder depth / heads & 12 / 6 & 12 / 6 \\
Predictor depth / heads & 4 / 4 & 4 / 4 \\
Projection/Predictor dim & 128 & 128 \\
Patch size & 25 samples & 25 samples \\
Channel Queries & 16 & 16 \\
Channel Mixer Dim & 32 & 32 \\
\midrule
\multicolumn{3}{l}{\textbf{Masking and Objective}} \\
Mask ratio & 0.6 & 0.6 \\
Mask block sizes & 5--10 patches & 5--10 patches \\
Global crops & 1 $\times$ 16 s & 1 $\times$ 16 s \\
\midrule
\multicolumn{3}{l}{\textbf{Regularization}} \\
SigReg weight & 0.05 & 0.02 \\
Query loss weight & 1.0 & 1.0 \\
\midrule
\multicolumn{3}{l}{\textbf{Optimization}} \\
Learning rate & $1\times10^{-4}$ & $1\times10^{-4}$ \\
Weight decay & 0.05 & 0.05 \\
LR schedule & Warmup + cosine & Warmup + cosine \\
Warmup & 1K steps & 1k steps \\
Min LR & $1\times10^{-6}$ & $1\times10^{-6}$ \\
\midrule
\multicolumn{3}{l}{\textbf{Training}} \\
Max steps & 10k & 20k \\
Precision & bf16-mixed & bf16-mixed \\
\bottomrule
\end{tabular}
\end{table}

\subsection{Pretraining Algorithm}
\label{app:pretraining-algorithm}

\begin{algorithm}[H]
\caption{\ours Pretraining Step}
\label{alg:pretraining}
\begin{algorithmic}[1]
\REQUIRE Latent brain states $\mathbf{S} \in \mathbb{R}^{B \times N \times D}$ and channel-mixer affinity weights $\mathbf{W}$
\STATE Sample contiguous temporal mask $\mathbf{m} \subset \{1,\dots,N\}$
\STATE $\mathbf{Z} \leftarrow \texttt{Encoder}(\mathbf{S})$
\STATE $\mathbf{z}_{\mathrm{cls}} \leftarrow \texttt{Mean}(\mathbf{Z}, \mathrm{dim}=1)$
\STATE $\mathbf{T} \leftarrow \texttt{Projector}(\texttt{StopGrad}(\mathbf{Z}))$
\STATE $\mathbf{p}_{\mathrm{cls}} \leftarrow \texttt{Projector}(\mathbf{z}_{\mathrm{cls}})$
\STATE $\mathbf{Z}_{\mathrm{ctx}} \leftarrow \texttt{Encoder}(\mathbf{S}, \mathbf{m})$
\STATE $\mathbf{P}_{\mathrm{ctx}} \leftarrow \texttt{Projector}(\mathbf{Z}_{\mathrm{ctx}})$
\STATE $\hat{\mathbf{T}} \leftarrow \texttt{Predictor}(\mathbf{P}_{\mathrm{ctx}}, \mathbf{m})$
\STATE $\mathcal{L}_{\text{query}} \leftarrow \texttt{QueryLoss}(\mathbf{W})$
\STATE $\mathcal{L} \leftarrow \texttt{MSE}(\hat{\mathbf{T}}, \mathbf{T}[\mathbf{m}]) + \lambda \cdot \texttt{SIGReg}(\mathbf{p}_{\mathrm{cls}}) + \gamma \cdot \mathcal{L}_{\text{query}}$
\end{algorithmic}
\end{algorithm}

\section{StopGrad Ablation}
\label{app:stopgrad-ablation}

In Laya, StopGrad serves a narrower role than in I-JEPA or V-JEPA, where it is coupled with a momentum-updated EMA target encoder. Here, we use a single shared encoder with no separate teacher network; StopGrad only decouples the target representations from the predictor gradient path, while SIGReg prevents dimensional collapse.

\begin{table}[h]
\centering
\caption{StopGrad ablation under frozen linear probing. Removing StopGrad while keeping SIGReg fixed.}
\label{tab:stopgrad_ablation}
\begin{tabular}{llcc}
\toprule
Configuration & StopGrad & BCI Mean & Clinical Mean \\
\midrule
\ourss & \checkmark & \textbf{0.395} & \textbf{0.603} \\
\ourss & $\times$ & 0.391 & 0.559 \\
\oursf\ (20K) & \checkmark & 0.394 & 0.597 \\
\oursf\ (20K) & $\times$ & 0.353 & 0.466 \\
\bottomrule
\end{tabular}
\end{table}

At the \ourss scale (10K steps), removing StopGrad leaves BCI performance nearly unchanged ($-$0.4pp) but reduces clinical accuracy by 4.4pp. At 20K steps, the effect is substantially larger, with a 13.1pp clinical drop, indicating that StopGrad plays an increasingly important stabilizing role over longer training. Overall, SIGReg is the primary mechanism preventing collapse; StopGrad plays a complementary role in preserving representation quality during extended training.

\section{Pretraining Datasets}
\label{seec:pretraining-data}

\paragraph{TUH EEG.} We use the Temple University Hospital EEG corpus, a large-scale clinical dataset consisting of long-duration, multi-channel EEG recordings with substantial inter-subject and inter-session variability. We ensured that no subjects appearing in the TUH-based downstream validation or test sets were included during pretraining.~\citep{obeid2016tuh}

\paragraph{HBN EEG.}
We use EEG recordings from the Healthy Brain Network (HBN), a large-scale pediatric neurodevelopmental dataset comprising multimodal data collected from typically developing children and individuals with psychiatric or learning disorders under standardized experimental protocols ~\citep{shirazi2024hbn}.

\paragraph{NMT.}
We use the NMT Scalp EEG Dataset, a large-scale open-source corpus of clinically acquired EEG recordings labeled as normal or abnormal. The dataset comprises long-duration, multi-channel scalp EEG.~\citep{khan_nmt_2022}

\paragraph{MOABB.}
We use the following MOABB datasets for downstream evaluation: Shin2017A~\citep{shin2017a}, BNCI2014\_002~\citep{bnci2014_002}, Ofner2017~\citep{ofner2017}, GrosseWentrup2009~\citep{grossewentrup2009}, Stieger2021~\citep{stieger2021}, and Lee2019\_MI~\citep{lee2019mi}.

\paragraph{EEGDash.} We query EEGDash and use a diverse selection of available datasets \href{https://openneuro.org/datasets/ds003474}{ds003474} \href{https://openneuro.org/datasets/ds003478}{ds003478} \href{https://openneuro.org/datasets/ds003506}{ds003506} \href{https://openneuro.org/datasets/ds003518}{ds003518} \href{https://openneuro.org/datasets/ds003522}{ds003522} \href{https://openneuro.org/datasets/ds003523}{ds003523} \href{https://openneuro.org/datasets/ds003638}{ds003638} \href{https://openneuro.org/datasets/ds003655}{ds003655} \href{https://openneuro.org/datasets/ds003690}{ds003690} \href{https://openneuro.org/datasets/ds003838}{ds003838} \href{https://openneuro.org/datasets/ds004256}{ds004256} \href{https://openneuro.org/datasets/ds004279}{ds004279} \href{https://openneuro.org/datasets/ds004448}{ds004448} \href{https://openneuro.org/datasets/ds004504}{ds004504} \href{https://openneuro.org/datasets/ds004515}{ds004515} \href{https://openneuro.org/datasets/ds004532}{ds004532} \href{https://openneuro.org/datasets/ds004572}{ds004572} \href{https://openneuro.org/datasets/ds004579}{ds004579} \href{https://openneuro.org/datasets/ds004580}{ds004580} \href{https://openneuro.org/datasets/ds004584}{ds004584} \href{https://openneuro.org/datasets/ds004595}{ds004595} \href{https://openneuro.org/datasets/ds004602}{ds004602} \href{https://openneuro.org/datasets/ds004626}{ds004626} \href{https://openneuro.org/datasets/ds004718}{ds004718} \href{https://openneuro.org/datasets/ds004771}{ds004771} \href{https://openneuro.org/datasets/ds004883}{ds004883} \href{https://openneuro.org/datasets/ds004902}{ds004902} \href{https://openneuro.org/datasets/ds004942}{ds004942} \href{https://openneuro.org/datasets/ds005131}{ds005131} \href{https://openneuro.org/datasets/ds005305}{ds005305} \href{https://openneuro.org/datasets/ds005385}{ds005385} \href{https://openneuro.org/datasets/ds005410}{ds005410} \href{https://openneuro.org/datasets/ds005540}{ds005540} \href{https://openneuro.org/datasets/ds005863}{ds005863}
. \nocite{openneuro:ds003474,openneuro:ds003478,openneuro:ds003506,openneuro:ds003518,openneuro:ds003522,openneuro:ds003523,openneuro:ds003638,openneuro:ds003655,openneuro:ds004256,openneuro:ds004279,openneuro:ds004448,openneuro:ds004515,openneuro:ds004532,openneuro:ds004572,openneuro:ds004579,openneuro:ds004580,openneuro:ds004584,openneuro:ds004595,openneuro:ds004602,openneuro:ds004626,openneuro:ds004718,openneuro:ds004771,openneuro:ds004883,openneuro:ds004902,openneuro:ds004942,openneuro:ds005131,openneuro:ds005305,openneuro:ds005385,openneuro:ds005540,openneuro:ds005863}

\paragraph{Preprocessing.}
All recordings are resampled to a sampling rate of 250~Hz, band-pass filtered between 0.5--100~Hz, and notch filtered at both 50~Hz and 60~Hz.
Channel layouts are preserved in their native montages (10--20, BioSemi128, or 10--05), with reference channels removed where applicable.
To exclude non-informative segments, we apply leading-edge trimming based on signal statistics, discarding windows with near-zero variance or excessive clipping.
Signals are robust scaled as in \citet{defossez2022}.
Preprocessed recordings are then segmented into non-overlapping 120-second chunks and stored as WebDataset shards containing both signals and metadata.

\section{Benchmark Details}
\label{sec:benchmark-desc}

To ensure comparability, each model follows a preprocessing pipeline aligned with its original design, as implemented in EEG-Bench. Across models, channels are harmonized to standard 10--20 montages where required; missing channels are handled via zero-padding or interpolation according to the model implementation. Signals are segmented into fixed-length windows, with window duration determined by each model architecture and task configuration.

\subsection{Downstream Datasets}
\label{sec:downstream-data}

\subsubsection{BCI Tasks.}
BCI tasks operate on short, trial-based recordings (typically 2--10~s) collected under controlled experimental paradigms, primarily motor imagery and event-related potential decoding.

\textbf{Left Hand vs Right Hand Motor Imagery (LH/RH MI).}
Binary classification of imagined left- versus right-hand movements. Evaluated on BCI Competition IV-2a and IV-2b \citep{Tangermann2012BCICompIV}, PhysioNet Motor Imagery \citep{Schalk2004BCI2000,Goldberger2000PhysioNet}, and multiple motor imagery datasets including Weibo2014~\citep{Yi2014Weibo}, Cho2017 \citep{Cho2017OpenBMI}, Liu2022\citep{Liu2024StrokeMI}, Schirrmeister2017\citep{Schirrmeister2017}, Zhou2016\citep{Zhou2016}, and Kaya2018 \citep{Kaya2018}.

\textbf{Right Hand vs Feet Motor Imagery (RH/Feet MI).}
Right-hand vs feet motor imagery (RH/Feet MI) is evaluated on Weibo2014~\citep{Yi2014Weibo}, PhysioNet Motor Movement/Imagery~\citep{Goldberger2000PhysioNet}, BCI Competition IV-2a~\citep{Tangermann2012BCICompIV}, Barachant2012~\citep{Barachant2012Riemannian}, Faller2012~\citep{faller2012}, Scherer2015~\citep{Scherer2015}, Schirrmeister2017~\citep{Schirrmeister2017}, Zhou2016~\citep{Zhou2016}, and Kaya2018~\citep{Kaya2018}.

\textbf{Four-Class Motor Imagery (4-Class MI).}
Multi-class motor imagery classification across left hand, right hand, feet, and tongue. Evaluated on BCI Competition IV-2a and Kaya2018.

\textbf{Five Fingers Motor Imagery.}
Fine-grained motor imagery classification of individual finger movements, evaluated on Kaya2018~\citep{Kaya2018}.

\subsubsection{Clinical Tasks.}
Clinical tasks operate on long-duration EEG recordings (minutes to hours) collected in diagnostic or monitoring settings. Tasks include both recording-level classification and segment-level event detection.

\textbf{Abnormal EEG Detection.}
Binary classification of EEG recordings as normal or abnormal. Evaluated on the Temple University Hospital Abnormal EEG Corpus (TUAB)~\citep{obeid2016tuh}.

\textbf{Epilepsy Detection.}
Binary classification of epileptic versus non-epileptic recordings. Evaluated on the Temple University Hospital Epilepsy Corpus (TUEP)~\citep{obeid2016tuh}.

\textbf{Seizure Detection.}
Per-segment seizure event detection in continuous EEG recordings. Evaluated on the CHB-MIT dataset~\citep{Shoeb2009Seizure}.

\textbf{Sleep Staging.}
Per-epoch sleep stage classification (W, N1, N2, N3/4, REM). Evaluated on Sleep-Telemetry (Sleep-EDF)~\citep{Mourtazaev1995Sleep}.

\textbf{Artifact Detection.}
Binary and multi-class artifact classification tasks identifying EEG contamination (e.g., eye movement, muscle, electrode artifacts). Evaluated on the Temple University Artifact Corpus (TUAR)~\citep{Hamid2020TUAR}.

\textbf{Neurological and Psychiatric Disorder Classification.}
Recording-level classification tasks distinguishing patients from healthy controls, including Parkinson's disease~\citep{Cavanagh2018PD,Brown2020PD,Singh2018PD,Singh2020PD,Singh2021PD}, schizophrenia~\citep{Albrecht2019SCZ}, mild traumatic brain injury~\citep{Cavanagh2019mTBI}, and obsessive--compulsive disorder~\citep{Grundler2009OCD}. These tasks are evaluated on EEG datasets released via OpenNeuro and related repositories.

\paragraph{Segment-Based and Multi-Label Tasks.}
For seizure detection, sleep staging, and artifact detection, recordings are divided into 16-second windows. For each window, models predict 16 class labels (one per second), and F1 score is computed across all per-second predictions.

\section{Additional Analyses}

\subsection{\ours Dynamic Channel Mixer}
\begin{figure}[t]
\centering
\includegraphics[width=0.62\linewidth]{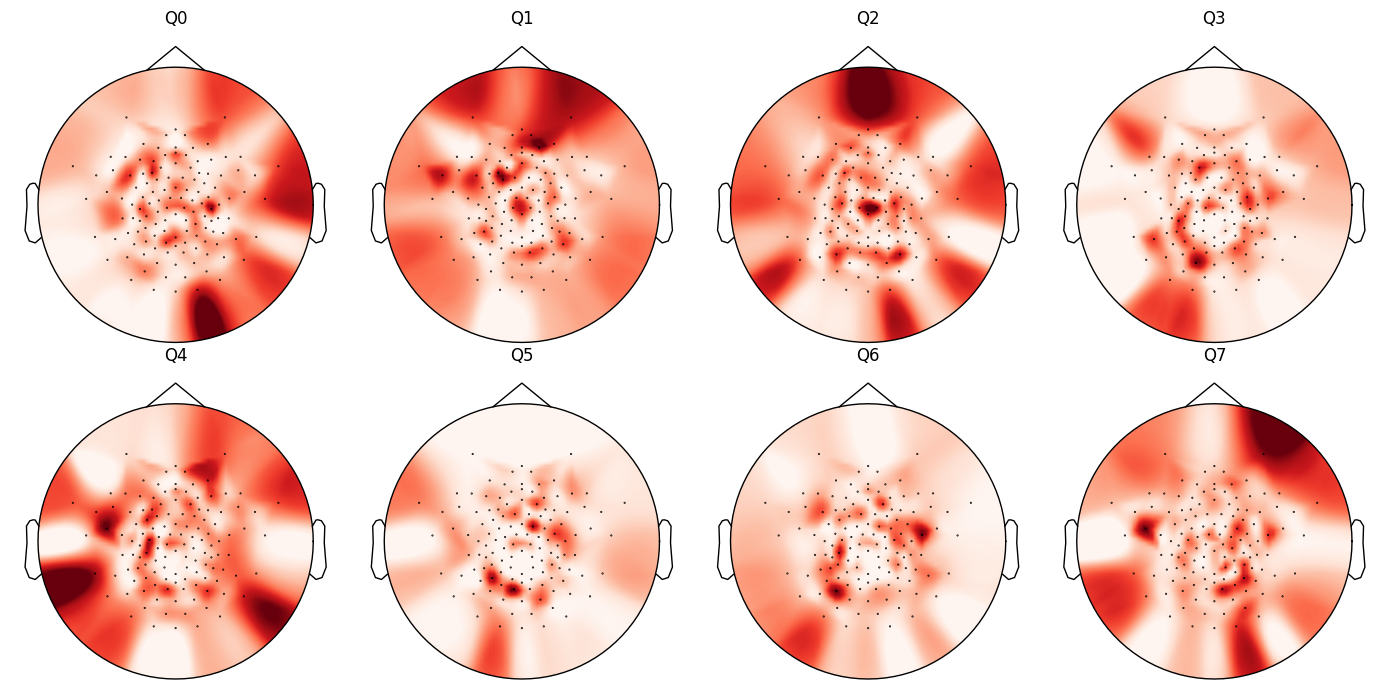}
\caption{Topoplot of the channel weights for eight queries on the HBN dataset during movie watching task (DiaryOfAWimpyKid).}
\label{fig:topoplot-hbn}
\end{figure}

\begin{figure}[t]
\centering
\includegraphics[width=0.62\linewidth]{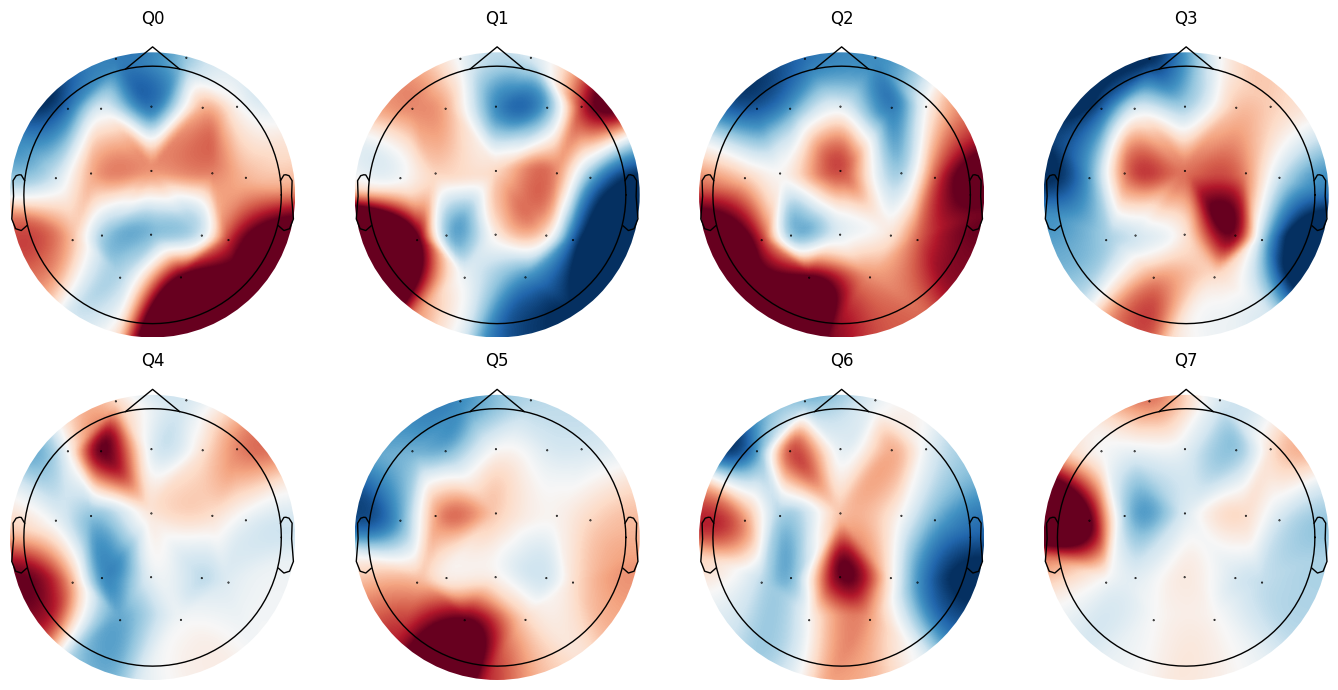}
\caption{Topoplot of the channel weights for eight queries on the TUH dataset.}
\label{fig:topoplot-tuh}
\end{figure}

As in LUNA, \ours learns to attend to different channels, each query attending to, and learning a rich latent brain state at each time step. In \cref{fig:topoplot-tuh} we show 8 queries and the channel weights they learn for one sample in the TUH dataset. In \cref{fig:topoplot-hbn} we show 8 queries and channel weights they learn for one sample in the HBN dataset during a movie watching task.

\subsection{Detailed Noise Robustness Results}
\label{app:noise-robust}

\begin{figure}[h]
\centering
\includegraphics[width=0.9\linewidth]{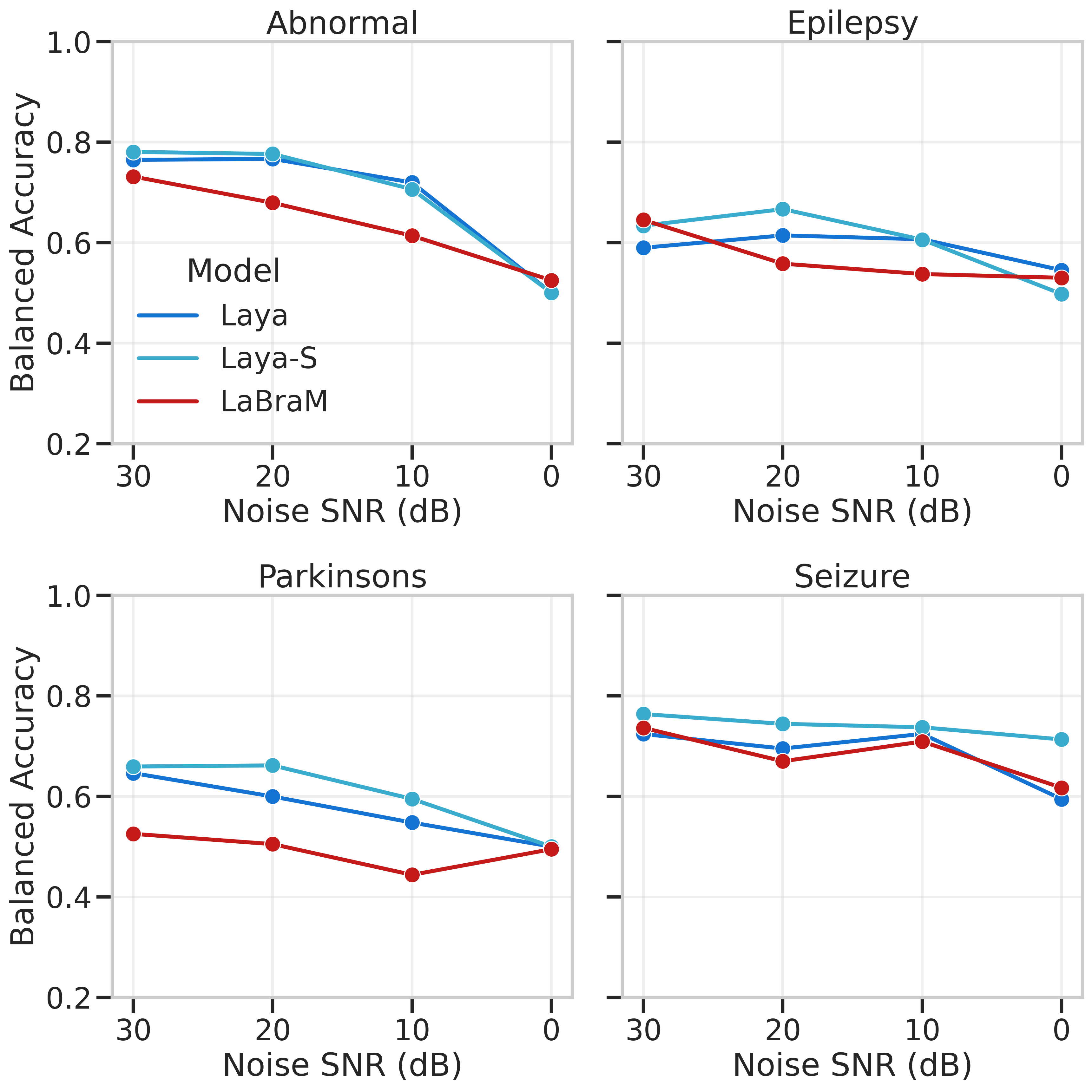}
\caption{\ours is resilient to noise across clinical tasks. Performance shown under combined noise (Gaussian, 1/f, EMG, channel dropout) at varying SNR levels.}
\label{noise-curves}
\end{figure}

\begin{figure}[p]
\centering
\includegraphics[width=0.9\linewidth]{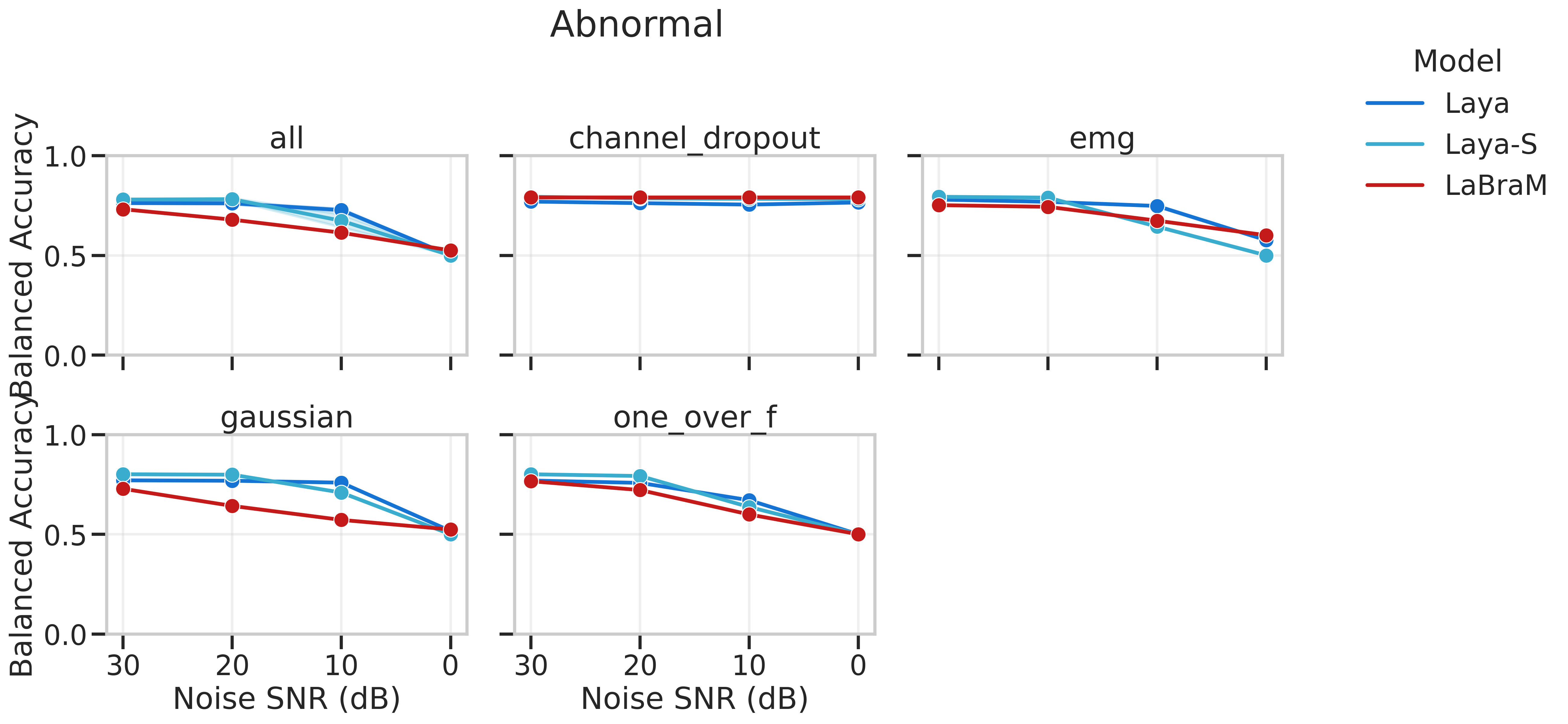}
\caption{Noise robustness on Abnormal EEG detection.}
\end{figure}

\begin{figure}[p]
\centering
\includegraphics[width=0.9\linewidth]{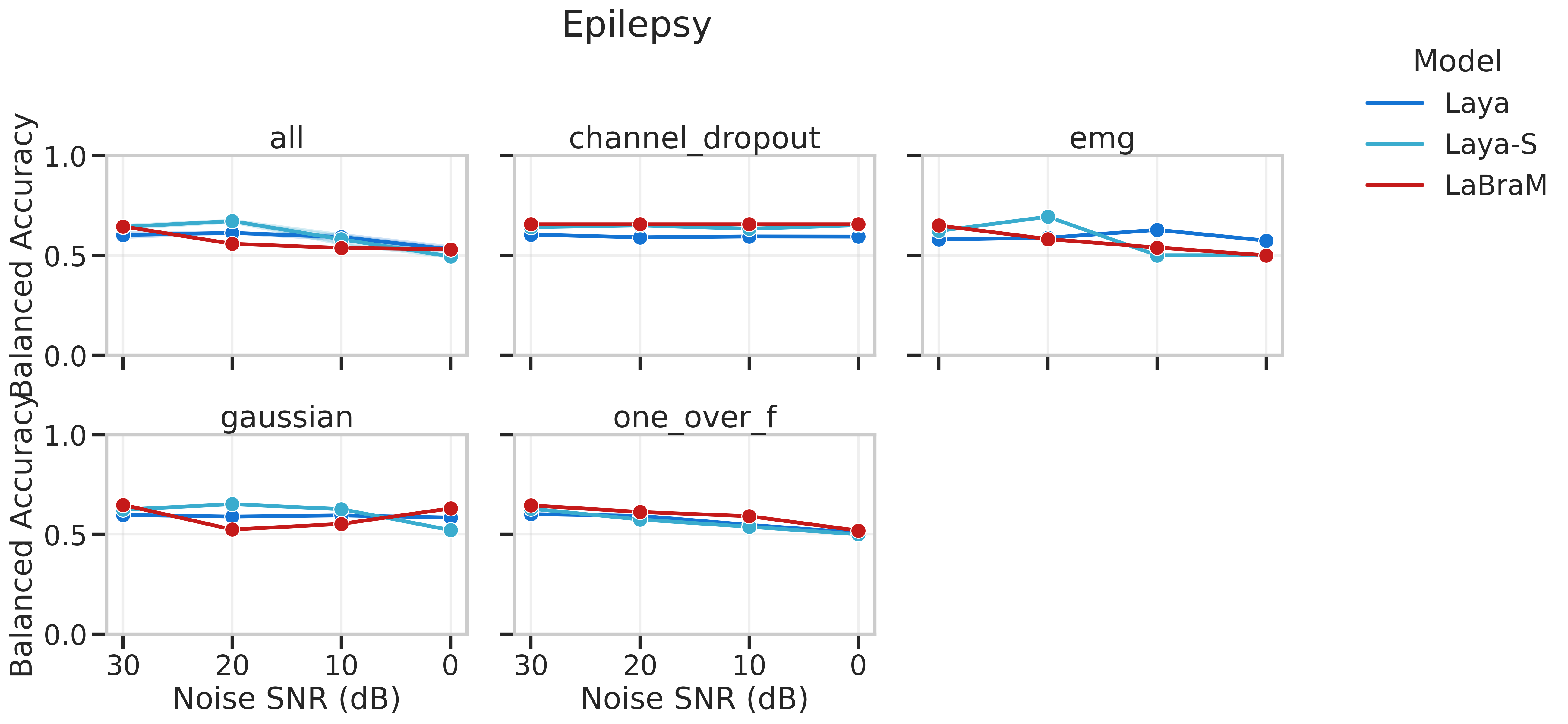}
\caption{Noise robustness on Epilepsy detection.}
\end{figure}

\begin{figure}[p]
\centering
\includegraphics[width=0.9\linewidth]{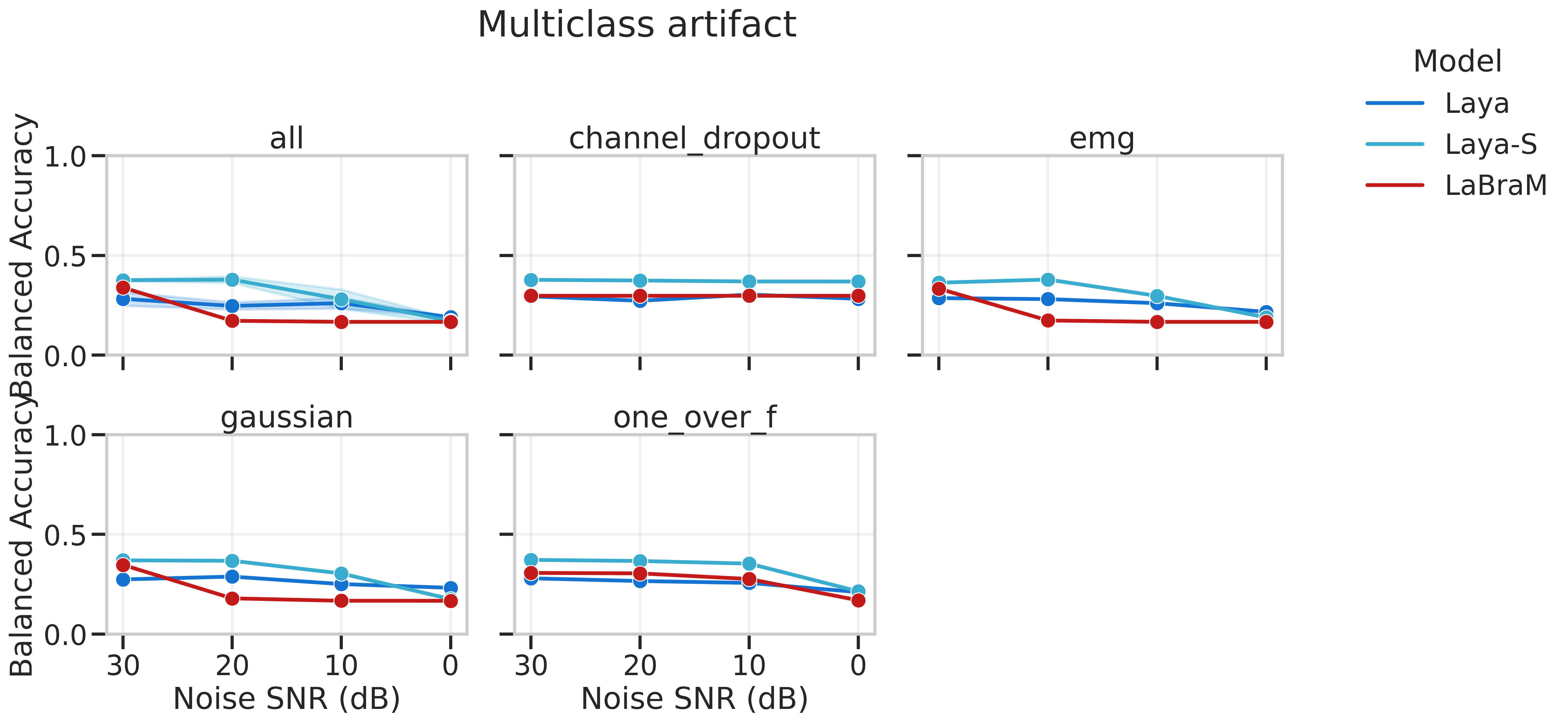}
\caption{Noise robustness on Multiclass Artifact classification.}
\end{figure}

\begin{figure}[p]
\centering
\includegraphics[width=0.9\linewidth]{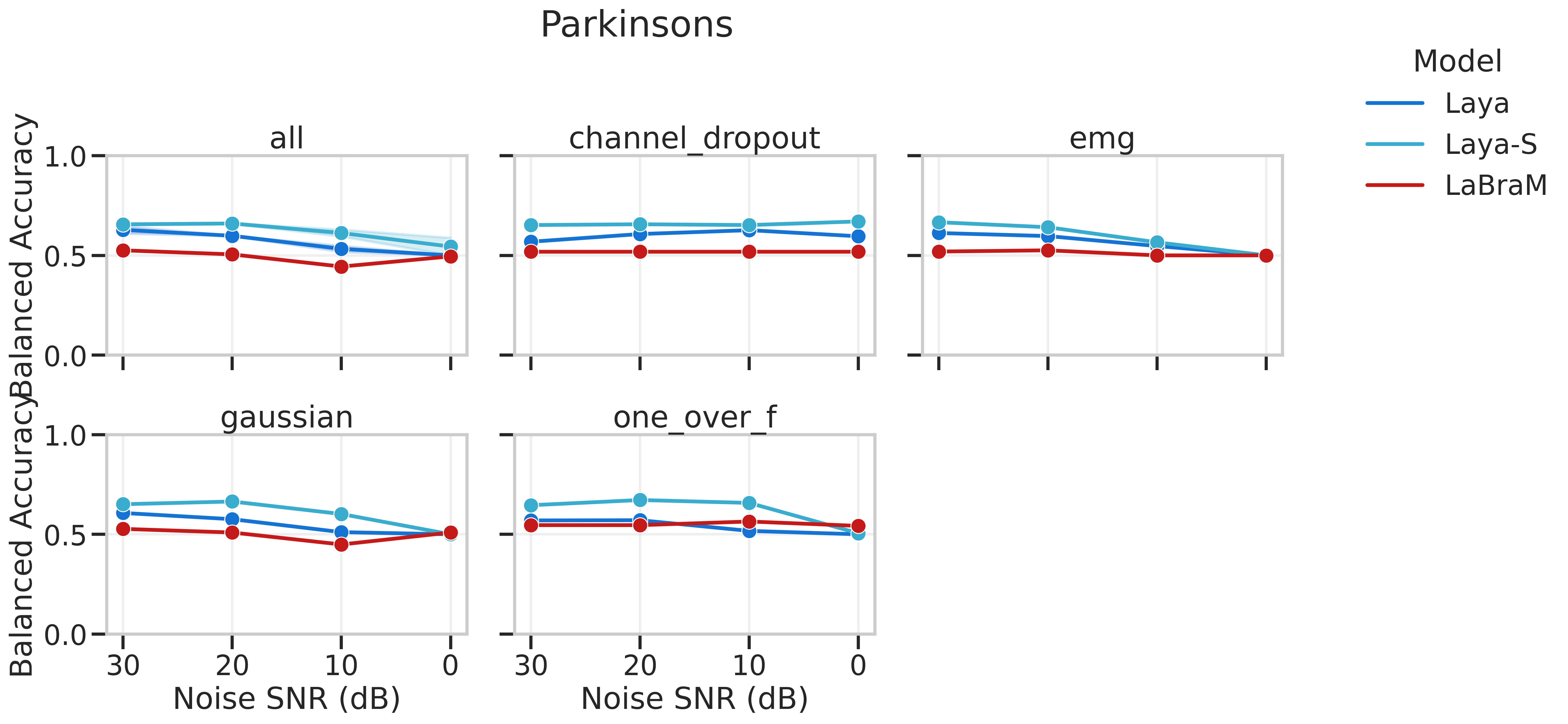}
\caption{Noise robustness on Parkinson's disease detection.}
\end{figure}

\begin{figure}[p]
\centering
\includegraphics[width=0.9\linewidth]{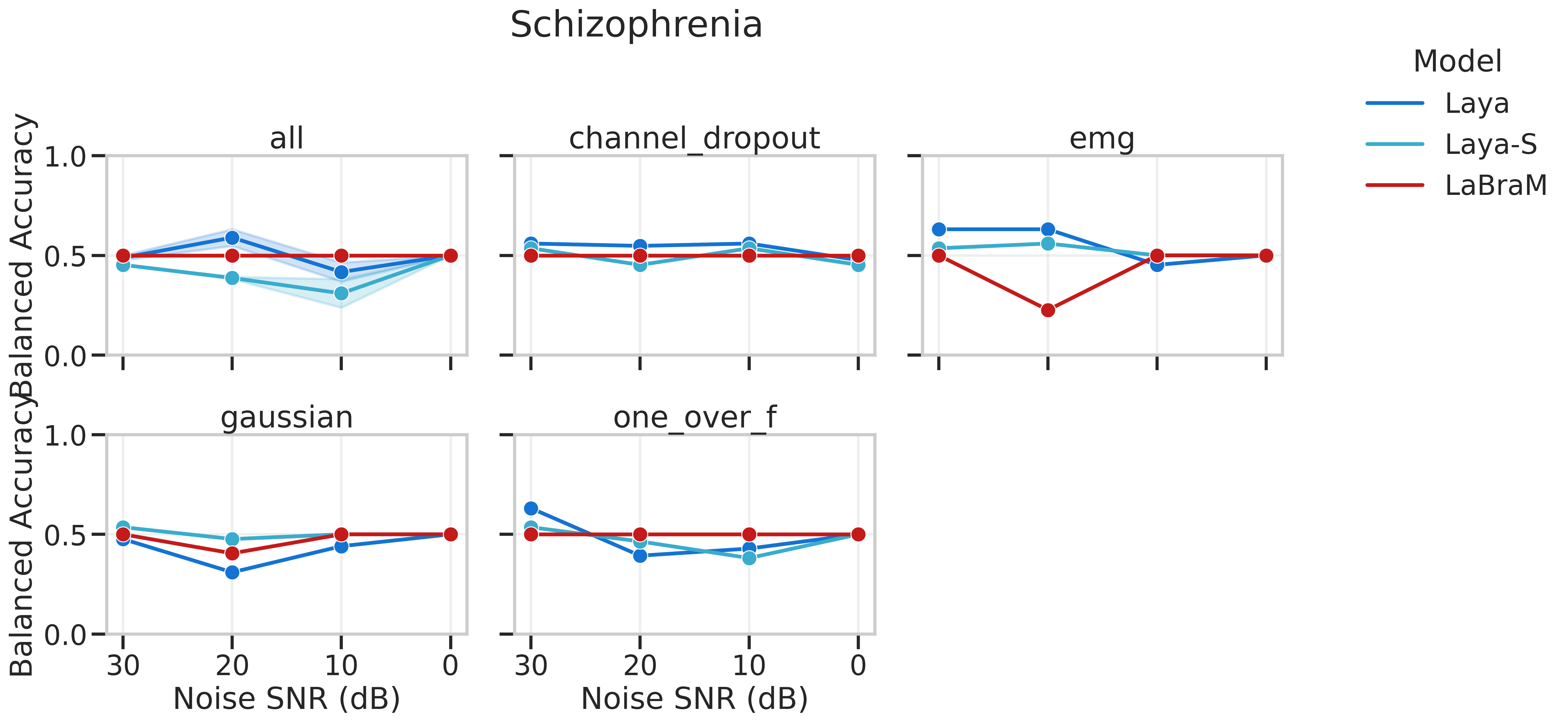}
\caption{Noise robustness on Schizophrenia detection.}
\end{figure}

\begin{figure}[p]
\centering
\includegraphics[width=0.9\linewidth]{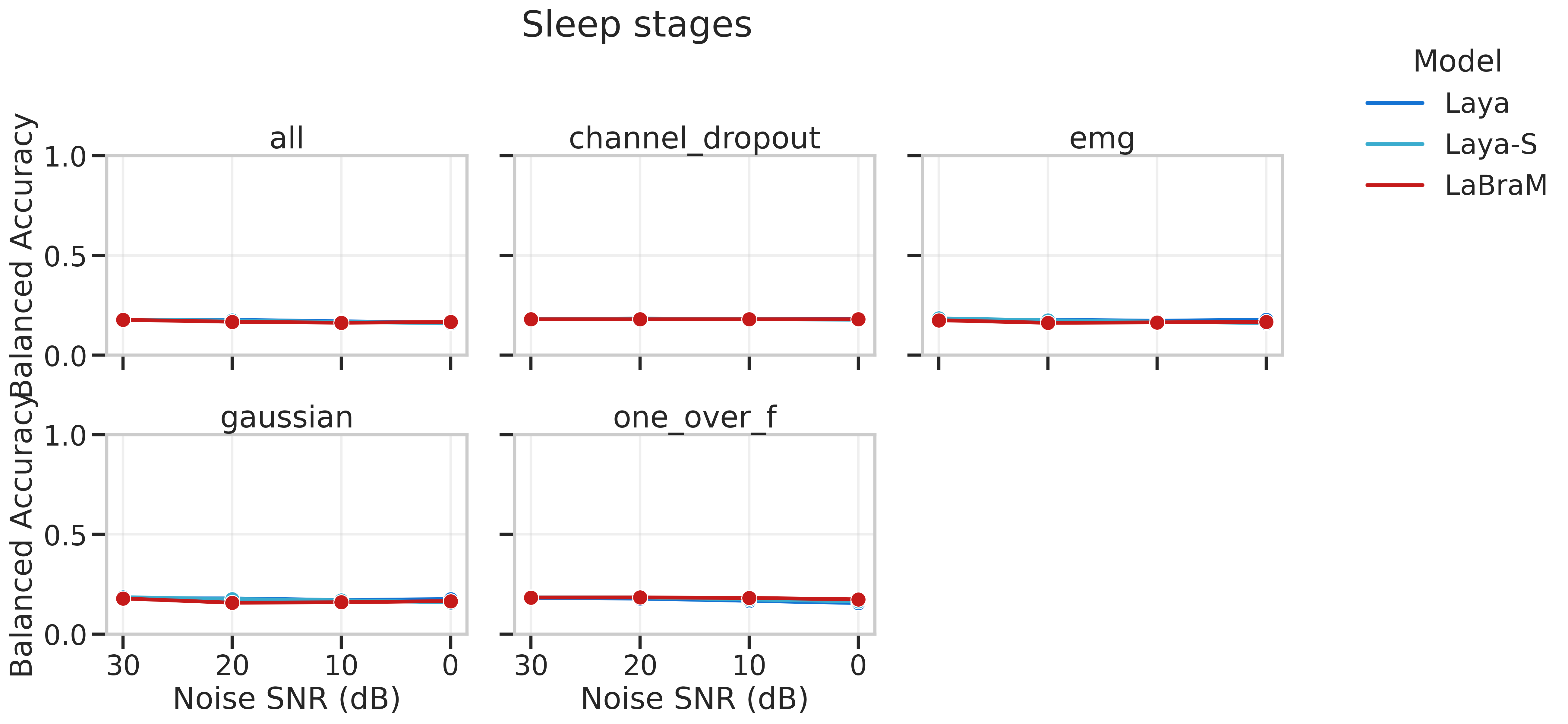}
\caption{Noise robustness on Sleep Staging.}
\end{figure}

\clearpage

\subsection{Further Examples of \ours Rich Temporal Embeddings on Seizure Recordings}
\label{app:pca}
\begin{figure}[p]
\centering
\includegraphics[width=0.95\linewidth]{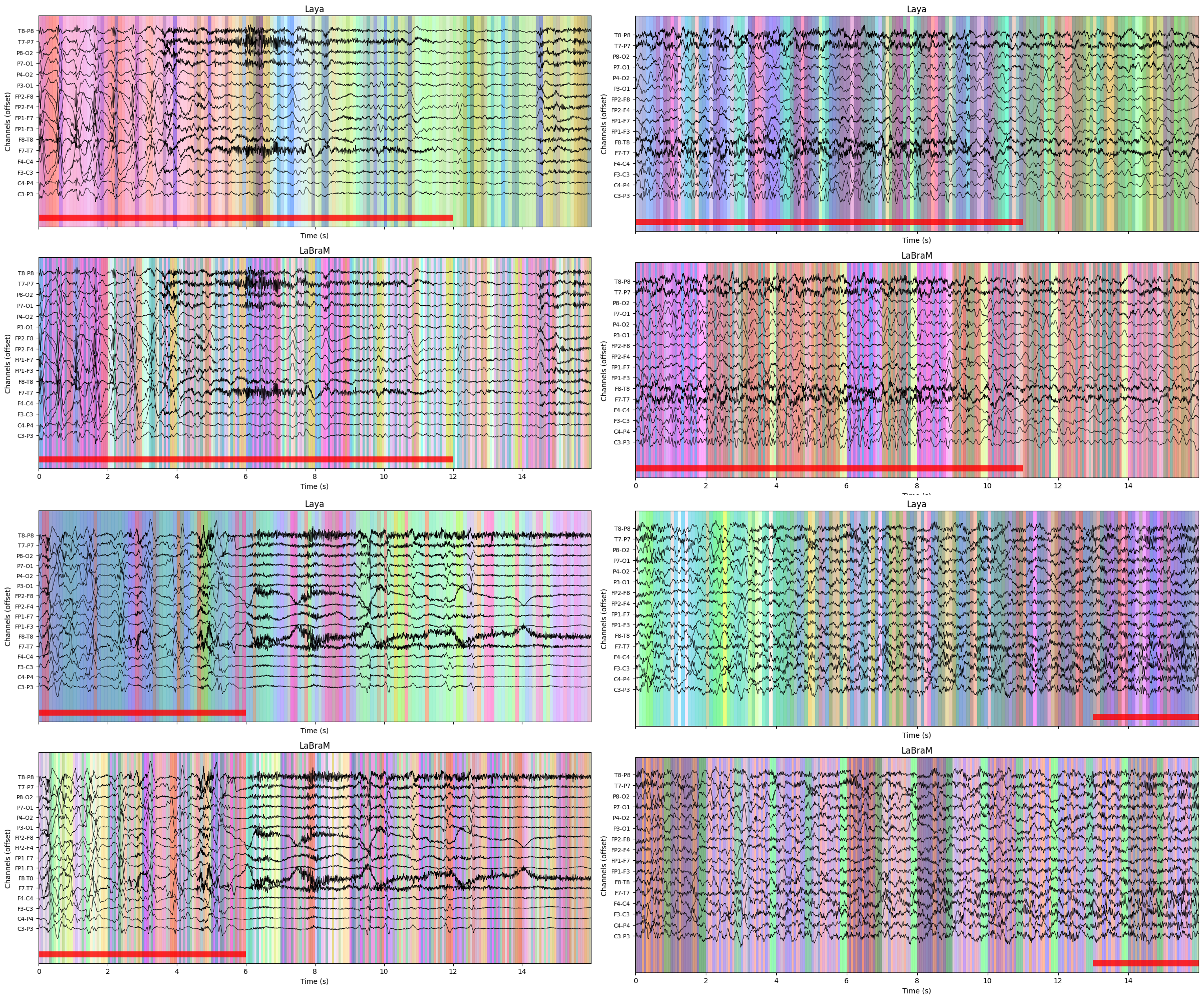}
\caption{
PCA visualization of patch embeddings on several seizure recordings. We map three principal components to RGB. Top: \oursf. Bottom: LaBraM. Red bar indicates seizure.
}
\label{fig:app-pca}
\end{figure}

Further examples of \ours ability to distinguish semantic meaning from noisy EEG signals can be found in \cref{fig:app-pca}.

\clearpage

\section{Additional Related Work}
\label{app:additional-related-work}

\paragraph{I-JEPA.}
I-JEPA applies the JEPA architecture to images by training a predictor to infer the latent embeddings of masked image blocks from a visible context block using an encoder--predictor architecture with an EMA target network, enabling semantic prediction without pixel-level reconstruction~\citep{assran2023ijepa}.

\paragraph{V-JEPA.}
V-JEPA extends JEPA to video by tokenizing clips into spatiotemporal blocks and training a predictor to infer masked block embeddings from visible ones using an encoder--predictor with an EMA target, operating entirely in representation space rather than reconstructing pixels~\citep{bardes2024vjepa}.

\paragraph{LeJEPA.}
LeJEPA augments the JEPA latent-prediction objective with Sketched Isotropic Gaussian Regularization (SIGReg), which encourages encoder embeddings to match an isotropic Gaussian, a distribution shown to minimize expected downstream risk. SIGReg uses random 1D projections to enforce this distribution with linear complexity, removing reliance on momentum-updated EMA teachers. The combined objective yields stable, collapse-free pretraining with a single trade-off hyperparameter and robust performance across architectures and domains~\citep{balestriero2025lejepa}.

\paragraph{LaBraM.}
LaBraM was among the first large-scale EEG foundation models, pretraining a transformer on over 2,500 hours of data using masked token prediction. The model discretizes EEG patches via a neural tokenizer trained to reconstruct Fourier spectra, then predicts masked tokens over a flattened channel--time sequence~\citep{Jiang2024LaBraM}.

\paragraph{CBraMod.}
CBraMod introduces a criss-cross transformer that models spatial and temporal dependencies separately using parallel attention mechanisms, rather than jointly attending over all channel--time tokens. The model employs patch-based masked EEG reconstruction with an asymmetric conditional positional encoding that dynamically encodes spatial and temporal relationships, improving adaptability across datasets with varying formats and reference schemes~\citep{wang2025cbramod}.

\paragraph{REVE.}
REVE adopts a masked autoencoding objective with structured spatio-temporal block masking and introduces a 4D positional encoding derived directly from electrode 3D coordinates and temporal indices. REVE supports arbitrary electrode layouts and sequence lengths without fine-tuning positional priors~\citep{ElOuahidi2025REVE}.

\paragraph{LUNA.}
LUNA introduces an efficient, topology-agnostic EEG foundation model that handles variable electrode configurations by compressing channel-wise patch features into a fixed-size latent space via learned queries and cross-attention, enabling linear scaling with channel count. Pretrained on over 21,000 hours of EEG using masked patch reconstruction, LUNA achieves strong transfer performance while reducing computational cost by orders of magnitude~\citep{Doner2025LUNA}.

\paragraph{S-JEPA.}
S-JEPA employs spatial block masking over EEG channels, masking contiguous scalp regions to encourage learning of spatial dependencies and dynamic channel attention. Experiments across multiple BCI paradigms demonstrate improved cross-dataset transfer relative to reconstruction-based SSL~\citep{guetschel2024s-jepa}.

\paragraph{Brain-JEPA.}
Brain-JEPA is an fMRI foundation model that predicts latent representations of masked spatiotemporal regions. The model introduces functional gradient-based positional encodings and structured spatiotemporal masking to address the lack of a natural spatial ordering in brain data~\citep{dong2024brainjepa}.

\paragraph{EEG-VJEPA.}
EEG-VJEPA adapts Video-JEPA to EEG by treating multichannel recordings as video-like spatiotemporal sequences and performing large contiguous spatiotemporal block masking with latent-space prediction. The model employs a ViT backbone with EMA-stabilized target encoders to learn long-range temporal and cross-channel dependencies~\citep{hojjati2025videoeegadaptingjoint}.

\paragraph{EEG-DINO.}
EEG-DINO is a large-scale EEG foundation model based on hierarchical self-distillation rather than masking-based prediction. EEG-DINO is pretrained on approximately 9,000 hours of EEG but is evaluated on only three downstream tasks, two of which overlap with its pretraining corpus, limiting conclusions about representation generality. The method aligns representations across multiple channel- and time-augmented views using a teacher--student framework with decoupled spatial and temporal positional embeddings~\citep{wang2025eegdino}.

\section{Broader Impacts}
\label{app:broader-impacts}

This work studies self-supervised EEG representation learning. Potential benefits include more reusable EEG representations for clinical neuroscience and more sample-efficient downstream modeling when labeled data are limited.

The model is not a clinical diagnostic system and should not be used for medical decision making without task-specific validation, prospective testing, and appropriate regulatory review. Because EEG data vary across populations, devices, montages, and sites, downstream use should account for distribution shift, privacy, and subgroup performance where metadata are available.

\newpage
\section*{NeurIPS Paper Checklist}

\begin{enumerate}

\item {\bf Claims}
    \item[] Question: Do the main claims made in the abstract and introduction accurately reflect the paper's contributions and scope?
    \item[] Answer: \answerYes{}
    \item[] Justification: The abstract and introduction clearly state that gains are demonstrated under frozen linear probing on clinical EEG tasks. Controlled MAE ablations in Section~\ref{sec:eegbench} support the causal attribution to the pretraining objective rather than architecture or data.

\item {\bf Limitations}
    \item[] Question: Does the paper discuss the limitations of the work performed by the authors?
    \item[] Answer: \answerYes{}
    \item[] Justification: The paper includes a dedicated Limitations subsection (Section 5.1) discussing scaling dynamics, benchmark scope, linear probing as an evaluation protocol, and BCI performance across self-supervised methods.

\item {\bf Theory assumptions and proofs}
    \item[] Question: For each theoretical result, does the paper provide the full set of assumptions and a complete (and correct) proof?
    \item[] Answer: \answerNA{}
    \item[] Justification: The paper does not include theoretical results. It cites existing theoretical work on JEPA and SIGReg from prior publications.

    \item {\bf Experimental result reproducibility}
    \item[] Question: Does the paper fully disclose all the information needed to reproduce the main experimental results of the paper to the extent that it affects the main claims and/or conclusions of the paper (regardless of whether the code and data are provided or not)?
    \item[] Answer: \answerYes{}
    \item[] Justification: Architecture, hyperparameters, masking strategy, optimizer settings, and evaluation protocol are fully described in the main paper and \cref{sec:pretraining-details,sec:downstream-data}.

\item {\bf Open access to data and code}
    \item[] Question: Does the paper provide open access to the data and code, with sufficient instructions to faithfully reproduce the main experimental results, as described in supplemental material?
    \item[] Answer: \answerNo{}
    \item[] Justification: Code will be released after the review period. All pretraining and evaluation datasets are publicly available and cited in the paper.

\item {\bf Experimental setting/details}
    \item[] Question: Does the paper specify all the training and test details (e.g., data splits, hyperparameters, how they were chosen, type of optimizer) necessary to understand the results?
    \item[] Answer: \answerYes{}
    \item[] Justification: Training details including optimizer, batch size, learning rate schedule, masking parameters, and linear probing protocol are described in the main paper and \cref{sec:pretraining-details,sec:downstream-data}.

\item {\bf Experiment statistical significance}
    \item[] Question: Does the paper report error bars suitably and correctly defined or other appropriate information about the statistical significance of the experiments?
    \item[] Answer: \answerYes{}
    \item[] Justification: All downstream results are reported as mean $\pm$ std over 5 random seeds, as noted in the table captions in \cref{sec:eegbench}.

\item {\bf Experiments compute resources}
    \item[] Question: For each experiment, does the paper provide sufficient information on the computer resources (type of compute workers, memory, time of execution) needed to reproduce the experiments?
    \item[] Answer: \answerYes{}
    \item[] Justification: The paper specifies GPU type (NVIDIA L40S) and number of devices (2 GPUs for pretraining) in the main text, and training steps for each variant in \cref{sec:pretraining-details}. Laya-S trained for approximately 1 day and Laya-B for approximately 2 days. 
    
\item {\bf Code of ethics}
    \item[] Question: Does the research conducted in the paper conform, in every respect, with the NeurIPS Code of Ethics \url{https://neurips.cc/public/EthicsGuidelines}?
    \item[] Answer: \answerYes{}
    \item[] Justification: The research uses existing de-identified EEG datasets and does not involve any direct human subjects research, data collection, or model deployment.

\item {\bf Broader impacts}
    \item[] Question: Does the paper discuss both potential positive societal impacts and negative societal impacts of the work performed?
    \item[] Answer: \answerYes{}
    \item[] Justification: Potential positive impacts and deployment risks are discussed in \cref{app:broader-impacts}. The work is methodological and evaluated on publicly available datasets; it does not constitute a clinical system or diagnostic tool.
    
\item {\bf Safeguards}
    \item[] Question: Does the paper describe safeguards that have been put in place for responsible release of data or models that have a high risk for misuse (e.g., pre-trained language models, image generators, or scraped datasets)?
    \item[] Answer: \answerNA{}
    \item[] Justification: This paper does not introduce any risk. The model processes de-identified clinical EEG data and is not applicable to harmful use cases.

\item {\bf Licenses for existing assets}
    \item[] Question: Are the creators or original owners of assets (e.g., code, data, models), used in the paper, properly credited and are the license and terms of use explicitly mentioned and properly respected?
    \item[] Answer: \answerYes{}
    \item[] Justification: All datasets and baseline models are cited with their original publications. All datasets used are publicly available for academic research use.

\item {\bf New assets}
    \item[] Question: Are new assets introduced in the paper well documented and is the documentation provided alongside the assets?
    \item[] Answer: \answerNA{}
    \item[] Justification: No new assets are released at submission time. Code and model checkpoints will be released after the review period.

\item {\bf Crowdsourcing and research with human subjects}
    \item[] Question: For crowdsourcing experiments and research with human subjects, does the paper include the full text of instructions given to participants and screenshots, if applicable, as well as details about compensation (if any)?
    \item[] Answer: \answerNA{}
    \item[] Justification: The paper uses existing publicly available EEG datasets. No crowdsourcing or direct human subjects research was conducted.

\item {\bf Institutional review board (IRB) approvals or equivalent for research with human subjects}
    \item[] Question: Does the paper describe potential risks incurred by study participants, whether such risks were disclosed to the subjects, and whether Institutional Review Board (IRB) approvals (or an equivalent approval/review based on the requirements of your country or institution) were obtained?
    \item[] Answer: \answerNA{}
    \item[] Justification: All data is sourced from pre-existing publicly available datasets, each collected under their own institutional approvals as documented in their original publications.

\item {\bf Declaration of LLM usage}
    \item[] Question: Does the paper describe the usage of LLMs if it is an important, original, or non-standard component of the core methods in this research? Note that if the LLM is used only for writing, editing, or formatting purposes and does \emph{not} impact the core methodology, scientific rigor, or originality of the research, declaration is not required.
    \item[] Answer: \answerNA{}
    \item[] Justification: LLMs were not used as part of the core methodology. Any use was limited to writing assistance.

\end{enumerate}

\end{document}